\documentclass{article}
\usepackage{fancyhdr}
\usepackage{authblk}
\usepackage[utf8]{inputenc}
\usepackage{threeparttable}
\usepackage{main}
\usepackage{microtype}
\usepackage{graphicx}
\newcommand{\starfull}{\includegraphics[height=1.5ex]{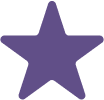}}
\newcommand{\starhalf}{\includegraphics[height=1.5ex]{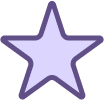}}
\usepackage[export]{adjustbox}
\usepackage{tabularx, booktabs}
\usepackage{times}
\usepackage{latexsym}
\usepackage{float}
\usepackage{amsmath}
\usepackage{colortbl}
\usepackage{footnote}
\newcolumntype{C}{>{\centering\arraybackslash}X}
\usepackage{caption}
\usepackage{enumitem}
\usepackage[most, skins]{tcolorbox}  
\usepackage{soul} 
\usepackage{soul}
\setulcolor{orange}
\setul{0.5ex}{0.3ex} 
\usepackage{bm}
\usepackage{arydshln}
\usepackage{booktabs}
\usepackage{multicol}
\usepackage{multirow}
\newcolumntype{M}[1]{>{\centering\arraybackslash}m{#1}}
\usepackage{color} 
\usepackage{colortbl}
\usepackage{bbding}
\usepackage{makecell}
\usepackage{mathtools}
\usepackage{imakeidx}
\usepackage{longtable}
\usepackage{wrapfig}
\usepackage[most]{tcolorbox}
\usepackage{array}  
\usepackage{graphicx}
\usepackage{ragged2e}
\usepackage{rotating}
\makeindex
\PassOptionsToPackage{table,dvipsnames,svgnames}{xcolor}
\usepackage{xcolor}
\usepackage{booktabs, multirow, colortbl}
\usepackage{pgfplots}
\pgfplotsset{compat=1.18}
\usepackage{etoolbox}
\usepackage{tikz}
\usepackage{pgfkeys}

\definecolor{blue1}{RGB}{240,248,255}  
\definecolor{blue2}{RGB}{218,236,255}  
\definecolor{red1}{RGB}{255,234,229}  
\definecolor{red2}{RGB}{255,216,204}  
\definecolor{lightmint}{RGB}{87,171,134} 
\definecolor{deeporange}{RGB}{247,159,127}

\usepackage{arydshln}
\usepackage{lipsum}
\usepackage{natbib}
\usepackage[toc]{multitoc}
\usepackage[edges]{forest}
\usepackage[normalem]{ulem}
\definecolor{mydarkblue}{rgb}{0,0.08,0.45}
\usepackage[colorlinks=true,linkcolor=mydarkblue,citecolor=mydarkblue,filecolor=mydarkblue,urlcolor=mydarkblue]{hyperref}
\usepackage{CJKutf8}
\usepackage{awesomebox} 
\usepackage{bbding}
\usepackage{booktabs}
\usepackage{geometry}
\definecolor{wkblue}{rgb}{0.2, 0.3, 0.6}
\definecolor{meta-color}{rgb}{0.5, 0.5, 0.5}
\definecolor{lightorange}{RGB}{226,104,58}
\definecolor{orange}{RGB}{255,184,159}
\definecolor{lightred}{RGB}{240, 128, 128}
\definecolor{custompurple}{HTML}{BCB2FF}
\definecolor{customdeepPurple}{HTML}{9A85FF}
\definecolor{shadowpurple}{HTML}{CBC4E7}
\definecolor{lightblue}{RGB}{220,230,250}
\usepackage{enumitem}
\usepackage{lscape} 
\usepackage{booktabs}
\usepackage{algorithm}  
\usepackage{subfig}
\usepackage{algorithmicx}
\usepackage{algpseudocode}
\usepackage{tabularx,booktabs}
\usepackage{makecell}
\usepackage{amssymb}
\usepackage{amsfonts}
\usepackage{wasysym}
\usepackage{pifont}  

\usepackage{hyperref}

\usepackage[tikz]{bclogo}
\usepackage[framemethod=tikz]{mdframed}
\definecolor{bgblue}{RGB}{245,243,253}
\definecolor{ttblue}{RGB}{91,194,224}

\mdfdefinestyle{mystyle}{%
  rightline=true,
  innerleftmargin=10,
  innerrightmargin=10,
  outerlinewidth=3pt,
  topline=false,
  rightline=true,
  bottomline=false,
  skipabove=\topsep,
  skipbelow=\topsep
}

\newtcolorbox{myboxi}[1][]{
  breakable,
  title=#1,
  colback=red!5,
  colbacktitle=red!5,
  coltitle=black,
  fonttitle=\bfseries,
  bottomrule=0pt,
  toprule=0pt,
  leftrule=2pt,
  rightrule=2pt,
  titlerule=0pt,
  arc=0pt,
  outer arc=0pt,
  colframe=red,
}

\newtcolorbox{myboxnote}[1][]{
  breakable,
  title=#1,
  colback=orange!0,
  colbacktitle=orange!0,
  coltitle=black,
  fonttitle=\bfseries,
  bottomrule=0pt,
  toprule=0pt,
  leftrule=2pt,
  rightrule=2pt,
  titlerule=0pt,
  arc=0pt,
  outer arc=0pt,
  colframe=orange,
}

\newtcolorbox{myboxii}[1][]{
  breakable,
  freelance,
  title=#1,
  colback=white,
  colbacktitle=white,
  coltitle=black,
  fonttitle=\bfseries,
  bottomrule=0pt,
  boxrule=0pt,
  colframe=white,
  overlay unbroken and first={
  \draw[red!75!black,line width=3pt]
    ([xshift=5pt]frame.north west) -- 
    (frame.north west) -- 
    (frame.south west);
  \draw[red!75!black,line width=3pt]
    ([xshift=-5pt]frame.north east) -- 
    (frame.north east) -- 
    (frame.south east);
  },
  overlay unbroken app={
  \draw[red!75!black,line width=3pt,line cap=rect]
    (frame.south west) -- 
    ([xshift=5pt]frame.south west);
  \draw[red!75!black,line width=3pt,line cap=rect]
    (frame.south east) -- 
    ([xshift=-5pt]frame.south east);
  },
  overlay middle and last={
  \draw[red!75!black,line width=3pt]
    (frame.north west) -- 
    (frame.south west);
  \draw[red!75!black,line width=3pt]
    (frame.north east) -- 
    (frame.south east);
  },
  overlay last app={
  \draw[red!75!black,line width=3pt,line cap=rect]
    (frame.south west) --
    ([xshift=5pt]frame.south west);
  \draw[red!75!black,line width=3pt,line cap=rect]
    (frame.south east) --
    ([xshift=-5pt]frame.south east);
  },
}

\usepackage{fancyhdr} 
\usepackage{blindtext} 
\usepackage{makecell}

\DeclareCaptionFont{black}{\color{black}}

\definecolor{opensiiBlue}{RGB}{127,126,236}
\definecolor{myblue}{rgb}{0.9, 0.1, 0.94}
\definecolor{mygreen}{rgb}{0.604, 0.522, 1.000}
\definecolor{myyellow}{rgb}{0.68, 0.6, 0.1}
\definecolor{fancygreen}{rgb}{0.33, 0.68, 0.20}
\definecolor{salmon}{rgb}{0.94, 0.52, 0.49}
\definecolor{tablegreen}{rgb}{0.82, 0.94, 0.75}
\definecolor{tableblue}{rgb}{0.81, 0.90, 0.94}
\definecolor{tablered}{rgb}{0.97, 0.85, 0.85}
\definecolor{tableorange}{rgb}{0.96, 0.85, 0.81}

\newenvironment{itemize*}%
 {\leftmargini=10pt\begin{itemize}%
  \setlength{\itemsep}{0pt}%
  \setlength{\parskip}{0pt}%
  }%
 {\end{itemize}}
\newenvironment{enumerate*}%
 {\begin{enumerate}%
  \setlength{\itemsep}{0pt}%
  \setlength{\parskip}{0pt}}%
 {\end{enumerate}}

\usepackage{listings}

\newcommand\JSONnumbervaluestyle{\color{blue}}
\newcommand\JSONstringvaluestyle{\color{red}}

\newif\ifcolonfoundonthisline

\makeatletter

\lstdefinestyle{json}
{
  showstringspaces    = false,
  keywords            = {false,true},
  alsoletter          = 0123456789.,
  morestring          = [s]{"}{"},
  stringstyle         = \ifcolonfoundonthisline\JSONstringvaluestyle\fi,
  MoreSelectCharTable =%
    \lst@DefSaveDef{`:}\colon@json{\processColon@json},
  basicstyle          = \ttfamily,
  keywordstyle        = \ttfamily\bfseries,
}

\newcommand\processColon@json{%
  \colon@json%
  \ifnum\lst@mode=\lst@Pmode%
    \global\colonfoundonthislinetrue%
  \fi
}

\lst@AddToHook{Output}{%
  \ifcolonfoundonthisline%
    \ifnum\lst@mode=\lst@Pmode%
      \def\lst@thestyle{\JSONnumbervaluestyle}%
    \fi
  \fi
  \lsthk@DetectKeywords%
}

\lst@AddToHook{EOL}%
  {\global\colonfoundonthislinefalse}

\makeatother
\usepackage{etoolbox}
\usepackage{natbib}
\usepackage{url}
\newcounter{bibcount}
\makeatletter
\patchcmd{\@lbibitem}{\item[}{\item[\hfil\stepcounter{bibcount}{[\thebibcount]}}{}{}
\renewcommand\NAT@bibsetup%
  [1]{\setlength{\leftmargin}{\bibhang}\setlength{\itemindent}{-\parindent}%
      \setlength{\itemsep}{\bibsep}\setlength{\parsep}{\z@}}
\makeatother

\definecolor{mybrown}{RGB}{128,64,0}

\definecolor{titlecolor}{HTML}{4c9cff}

\begin{document}

\thispagestyle{fancy}
\fancyhead{}
\lhead{\includegraphics[height=0.8cm]{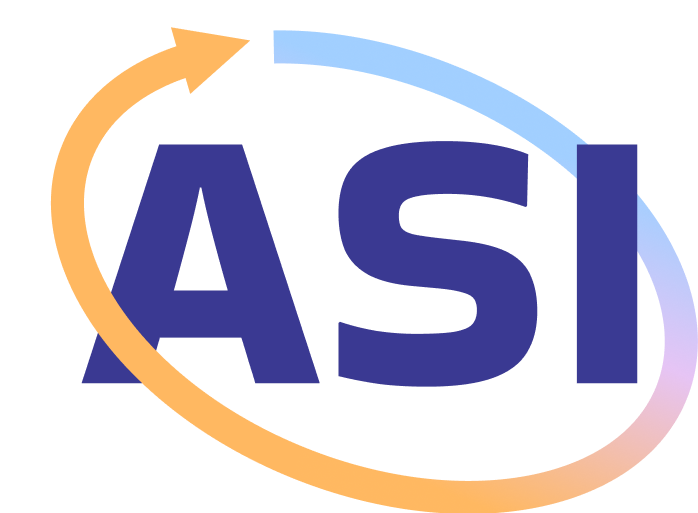}}
\rhead{%
  \raisebox{-0.1cm}{\includegraphics[height=0.85cm]{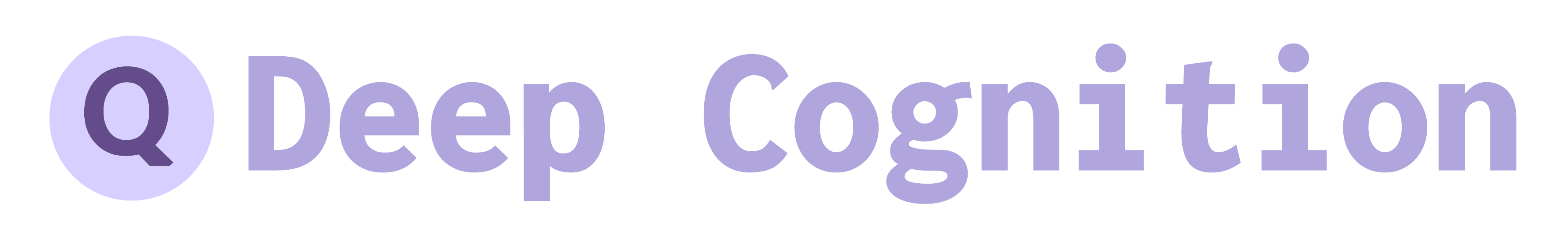}}%
}

\renewcommand{\headrulewidth}{0.8pt}
\setlength{\headsep}{5mm} 
\begin{center}
    \includegraphics[width=\textwidth]{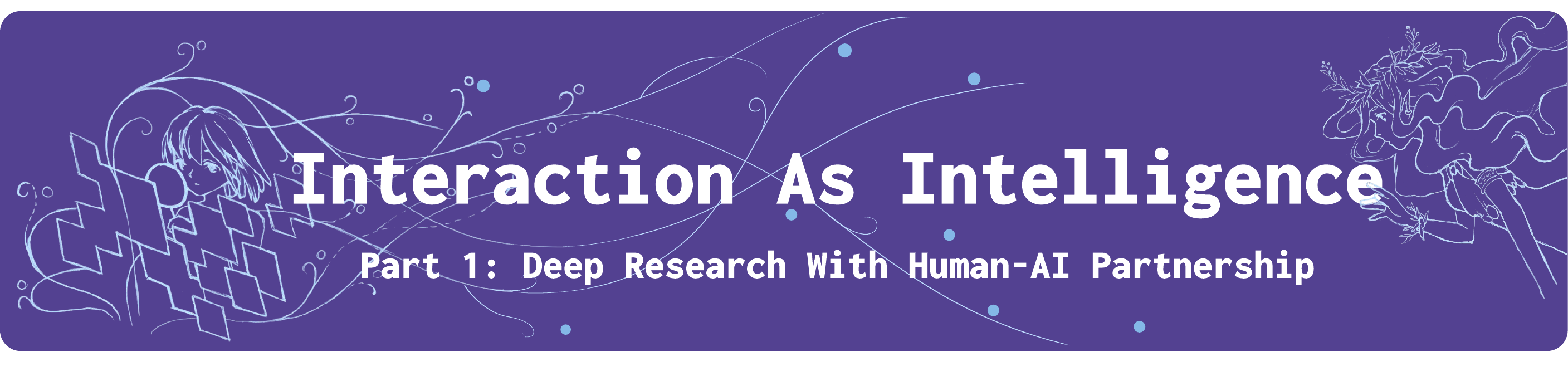} 
\end{center}
\vspace{-1.2em}

\begin{center}

\textit{
Lyumanshan Ye\textsuperscript{\tiny 2},
Xiaojie Cai\textsuperscript{\tiny 2},
Xinkai Wang\textsuperscript{\tiny 3,5},
Junfei Wang\textsuperscript{\tiny 4},
Xiangkun Hu\textsuperscript{\tiny 3},
Jiadi Su\textsuperscript{\tiny 4,5},
Yang Nan\textsuperscript{\tiny 4,5},
Sihan Wang\textsuperscript{\tiny 3},
Bohan Zhang\textsuperscript{\tiny 3},
Xiaoze Fan\textsuperscript{\tiny 3},
Jinbin Luo\textsuperscript{\tiny 3},
Yuxiang Zheng\textsuperscript{\tiny 5},
Tianze Xu\textsuperscript{\tiny 5},
Dayuan Fu\textsuperscript{\tiny 5},
Yunze Wu\textsuperscript{\tiny 5},
Pengrui Lu\textsuperscript{\tiny 5},
Zengzhi Wang\textsuperscript{\tiny 5},
Yiwei Qin\textsuperscript{\tiny 5},
Zhen Huang\textsuperscript{\tiny 5},
Yan Ma\textsuperscript{\tiny 5},
Zhulin Hu\textsuperscript{\tiny 5},
Haoyang Zou\textsuperscript{\tiny 5},
Tiantian Mi\textsuperscript{\tiny 5},
Yixin Ye\textsuperscript{\tiny 5},\\
Ethan Chern\textsuperscript{\tiny 5},
Pengfei Liu\textsuperscript{\tiny 1}
}\\

\footnotetext[1]{Supervision, corresponding author, email: pengfei@sjtu.edu.cn
\textsuperscript{2}Project Lead
\textsuperscript{3}Algorithm
\textsuperscript{4}UI Interface
\textsuperscript{5}Annotation}

\vspace{0.5em} 
\textcolor{opensiiBlue}{\textsc{SII-GAIR, SJTU}: \url{https://opensii.ai/}}

\end{center}

\vspace{8pt}

\begin{abstract}
\vspace{-0.5em}
This paper introduces ``Interaction as Intelligence'' research series, presenting a fundamental reconceptualization of human-AI relationships in deep research tasks. Traditional approaches treat interaction merely as an interface for accessing AI capabilities---a conduit between human intent and machine output. We propose that \textbf{interaction itself constitutes a fundamental dimension of intelligence}. As AI systems engage in extended thinking processes for complex research tasks, meaningful interaction transitions from an optional enhancement to an essential component of effective intelligence. Current deep research systems uniformly adopt an ``input-wait-output'' paradigm where users initiate queries and receive results after prolonged black-box processing. This approach leads to error cascade effects, inflexible research boundaries that prevent question refinement during investigation, and missed opportunities for expertise integration. To address these limitations, we introduce \textbf{Deep Cognition}, a system that transforms the human role from giving instructions to \textbf{cognitive oversight}---a mode of engagement where humans guide AI thinking processes through strategic intervention at critical junctures. Deep cognition implements three key innovations: (1) Transparent, controllable, and interruptible interaction that reveals AI reasoning and enables intervention at any point; (2) Fine-grained bidirectional dialogue; and (3) Shared cognitive context where the system observes and adapts to user behaviors without explicit instruction. User evaluation demonstrates that this cognitive oversight paradigm significantly outperforms the strongest baseline across six key metrics: \textbf{Transparency (+20.0\%), Fine-Grained Interaction (+29.2\%), Real-Time Intervention (+18.5\%), Ease of Collaboration (+27.7\%), Results-Worth-Effort (+8.8\%), and Interruptibility (+20.7\%)}. 
Evaluations on challenging deep research problems show \textbf{31.8\% to 50.0\% points of improvements} over competitive deep research systems. 

\vspace{10pt}  
\begin{figure}[htbp]
    \centering
    \includegraphics[width=1\linewidth]{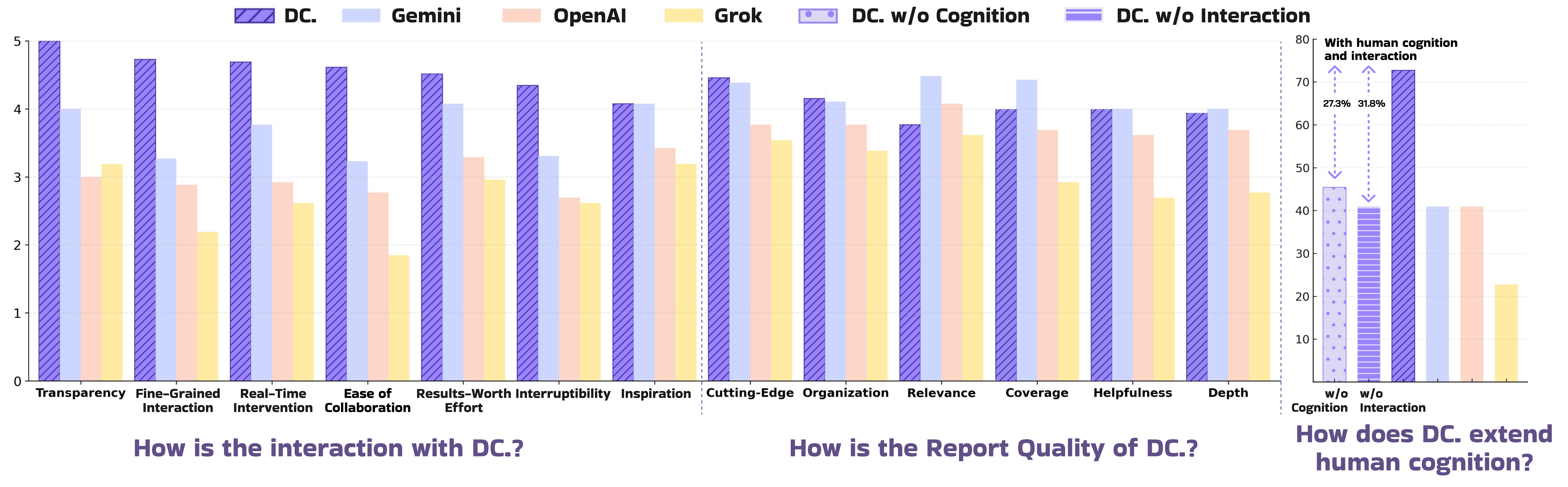}
    \caption{\textbf{Overall evaluation results.} We present the user evaluation (seven metrics on the left part), report quality (six metrics in the middle), and evaluation results on deep research problems (the right part) with three conditions: \textcolor{customdeepPurple!90!white}{\textit{Without Cognition}}, \textcolor{customdeepPurple!50!black}{\textit{With Cognition \& Interaction}}, and \textcolor{customdeepPurple!90!white}{\textit{Without Interaction}}. These results demonstrate the cognitive amplification effect of deep cognition when users collaborate with AI to perform long and complex tasks. ``DC.'' stands for ``deep cognition''.}
    \label{fig:fig2}
\end{figure}

\end{abstract}

\newpage

\pagestyle{fancy}
\lhead{\leftmark}
\renewcommand{\headrulewidth}{0.7pt}
\setlength{\headsep}{5mm}
\clearpage

\section{Introduction}
\label{sec:intro}


\begin{quote}
\emph{``Intelligence is not the property of an isolated mind, but emerges in the dance between minds. The question is not how smart the individual components are, but how brilliantly they interact.''} 
\end{quote}

\begin{figure}
    \centering
    \includegraphics[width=1\linewidth]{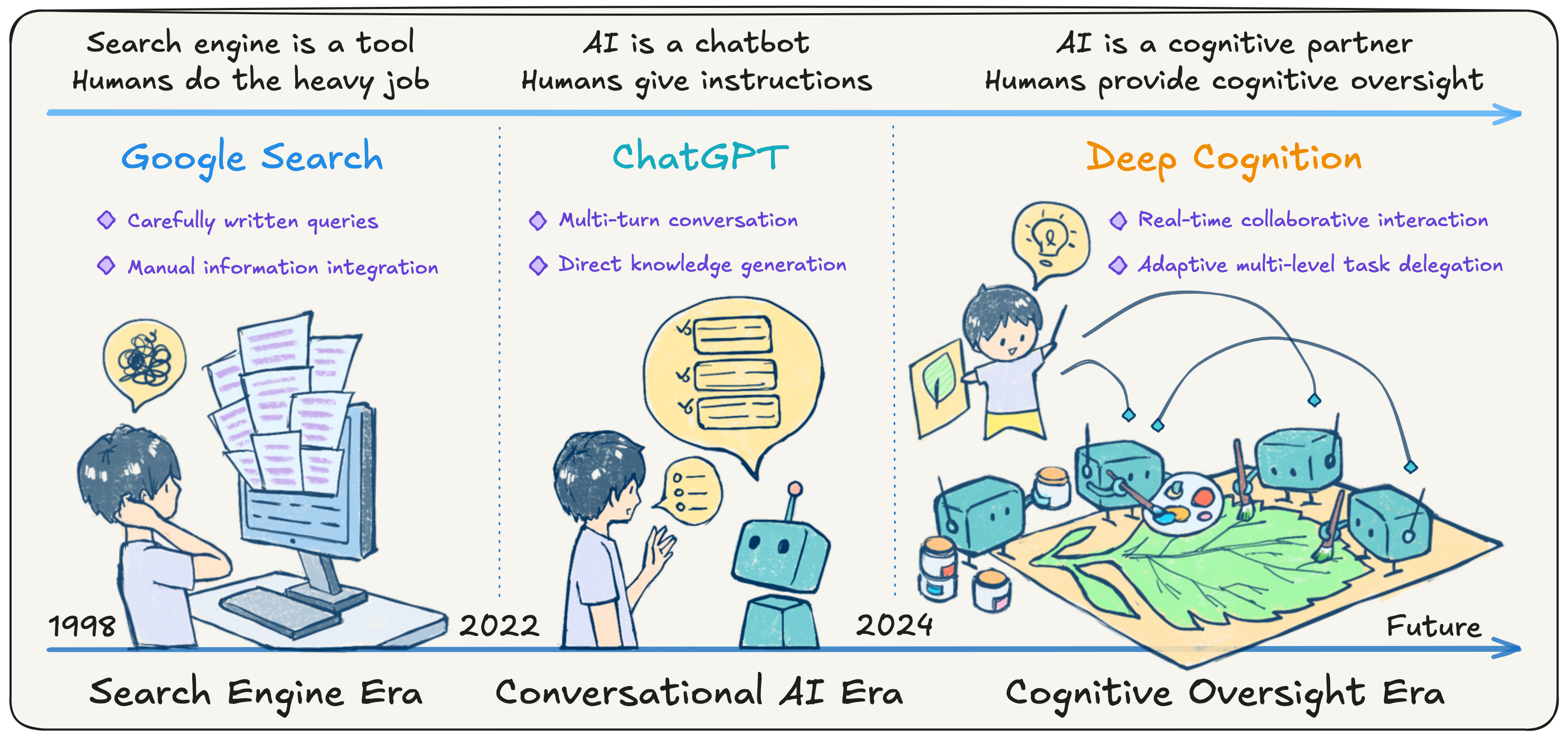}
    \caption{
    The evolution of human machine interaction from \textbf{operational interaction} (manual search) through \textbf{conversational interaction} (ChatGPT-style dialogue) to \textbf{cognitive interaction} (deep cognition). Our proposed paradigm transforms human-AI collaboration from periodic consultation to continuous cognitive partnership, where intelligence emerges through real-time interaction rather than autonomous processing.
    }
    \label{fig:enter-label-intro}
\end{figure}

As artificial intelligence (AI) capabilities have advanced dramatically through large language models (LLMs)~\cite{luo2024improve, radford2018improving, radford2021learningtransferablevisualmodels, brown2020languagemodelsfewshotlearners, brown2024largelanguagemonkeysscaling}, a fundamental question emerges: \textbf{\textit{How to build the equality relationship between human and machine intelligence in the age of AI?}} 
The prevailing trajectory in AI development has emphasized scaling model parameters~\cite{kaplan2020scalinglawsneurallanguage, 10.5555/3600270.3602446, DBLP:journals/tmlr/WeiTBRZBYBZMCHVLDF22}, expanding training data~\cite{yang2025qwen3technicalreport, metaai2025llama4}, and refining architectures~\cite{deepseekai2025deepseekv3technicalreport, minimax2025minimax01scalingfoundationmodels, poli2024mechanisticdesignscalinghybrid}---creating increasingly autonomous black boxes that assume minimal human input beyond simple prompting~\cite{liu2023pre,10.1145/3544548.3581001} or decision-making~\cite{yin2025bridging}. This pathway implicitly assumes that the ultimate form of artificial intelligence would require minimal human input, with interaction reduced to simple prompting or instruction~\cite{10.1145/3544548.3581001} or AI-assisted decision-making~\cite{yin2025bridging}. We contend that this assumption fundamentally mischaracterizes the nature of intelligence itself. This paradigm positions humans as external operators who provide initial prompts and consume final outputs while remaining excluded from the cognitive process itself, treating human intelligence as merely an instructor rather than a collaborative partner.
However, intelligence---whether human or artificial---is inherently interactive, contextual, and collaborative~\cite{hutchins1995cognition, 80e5fb31-5983-3e1c-b800-a54e6f555319, doi:10.1126/science.1193147}. The most sophisticated human thinking rarely occurs in isolation but emerges through dialogue, feedback, refinement, and the integration of diverse perspectives. Consider the nature of breakthrough scientific discoveries or complex problem-solving scenarios: They invariably involve iterative cycles of hypothesis formation, testing, revision, and collaborative refinement---demonstrating that expert consultation and interdisciplinary dialogue are integral components of intelligence at its highest levels.

As AI systems approach advanced cognitive capabilities powered by inference-time scaling~\cite{openai-o1}—enabling thought-level communication where strategic human oversight can leverage vast AI execution power~\cite{xia2025generative}---the need for meaningful interaction transforms and intensifies. This is especially critical for extended AI tasks~\cite{kwa2025measuringaiabilitycomplete} spanning hours to days, which fundamentally alter human-AI collaboration dynamics.

We propose a radical reconceptualization: \emph{interaction itself constitutes a fundamental dimension of intelligence, enabling capabilities that neither humans nor AI can achieve independently}. This recognition represents a pivotal shift in how we conceptualize AI progress: \textbf{\emph{from minimizing human involvement to optimizing human-AI cognitive partnership}}.
The evolution of human-computer interaction in information seeking exemplifies this fundamental transformation. As shown in Figure~\ref{fig:enter-label-intro}, initially, interaction was purely operational: users manually retrieved information through explicit queries and integrated fragmented results themselves. The conversational era, exemplified by LLM-based chatbots such as ChatGPT~\cite{openai2022chatgpt}, introduced natural language dialogue but still demanded extensive multi-turn conversations to accomplish complex tasks. We now stand at the threshold of a new paradigm where humans and AI communicate at the cognition level, enabling strategic intervention rather than constant guidance.

This transition is particularly evident in systems designed for {Deep Research} tasks~\cite{openai_deep_research_system_card, gemini_deep_research, perplexity_deep_research, zheng2025deepresearcher}---complex, extended cognitive processes involving dynamic information retrieval, filter, understanding, analysis and synthesis. Current state-of-the-art research systems have pioneered capabilities for multi-step web browsing, data analysis, and report generation. However, these systems uniformly adopt an ``Input-Wait-Output'' interaction paradigm where users initiate a query, wait through an extended ``black box'' processing period (typically 5-30 minutes), and eventually receive a comprehensive result.
This approach reflects the persistent assumption that interaction is merely a necessary cost rather than a source of value. Yet these systems fundamentally suffer from critical deficiencies: early errors~\cite{cemri2025multiagentllmsystemsfail} compound without correction, systems cannot adapt to evolving requirements, domain expertise remains inaccessible at crucial moments, and opaque processing prevents human-AI collaboration.

These deficiencies stem from a fundamental misalignment: systems that minimize human involvement during processing cannot address problems that require adaptive guidance and expert intervention~\cite{bainbridge1983ironies}. Rather than viewing interaction as overhead, we need systems that enable continuous and meaningful collaboration throughout the research process. To address this fundamental challenge, we develop~\textbf{deep cognition}---a systematic framework that transcends traditional automation by embedding real-time human expertise directly into AI reasoning processes for complex research tasks, guided by the following principles:

\begin{itemize*}
    \item \emph{\textbf{Cognitive Transparency:}} The system reveals its entire thinking process---from search strategies and query formulations to information evaluation and synthesis rationales---making AI cognition inspectable and editable at every stage. This transparency enables true thought-level interaction where humans can guide not just what the AI does, but also how it thinks.
    
    \item \emph{\textbf{Real-Time Intervention:}} Unlike conventional systems that operate in isolated processing cycles, deep cognition allows users to pause the research progress and input feedback and requirements at any moment. This creates continuous dialogue rather than discrete query-response cycles.

    \item \emph{\textbf{Fine-Grained Interaction:}} Users can engage with any specific element of the AI's output---questioning particular claims, requesting elaboration on specific points, or changing the research focus. This enables for targeted adjustments rather than starting over completely.
    
    \item \textbf{\textit{Adaptive Cognitive Context:}} Deep cognition evolves its research strategies by learning from accumulated interaction patterns, developing a sophisticated understanding of user preferences without explicit retraining. This adaptive capability is grounded in In-context Reinforcement Learning (ICRL), where neural networks learn algorithms purely through contextual conditioning~\cite{laskin2022incontextreinforcementlearningalgorithm, lin2024transformersdecisionmakersprovable, lee2023supervisedpretraininglearnincontext}. Just as ICRL models adapt and often surpass their training performance through context alone~\cite{huang2024incontextdecisiontransformerreinforcement, grigsby2024amagoscalableincontextreinforcement}, deep cognition progressively refines its collaborative approach, creating a dynamic partnership that better anticipates user needs over time.
    
\end{itemize*}

These principles fundamentally transform deep research from conventional question-and-answer exchanges into cognitive collaboration—what we term \textbf{cognitive oversight}. Rather than relegating humans to the role of passive tool operators, this framework establishes a synergistic reasoning process that harnesses the complementary strengths of human expertise and AI capabilities while mitigating their respective limitations. Through cognitive oversight, we move beyond the traditional paradigm of human-AI interaction toward a new form of augmented intelligence where strategic human insight and AI computational power merge into a unified cognitive system.

Through extensive experiments with real expert interactions, we demonstrate that deep cognition achieves substantial improvements or competitive over strongest baseline across all evaluation dimensions: Transparency (+20.0\%), Fine-Grained Interaction (+29.2\%), Real-Time Intervention (+18.5\%), Ease of Collaboration (+27.7\%), Results-Worth-Effort (+8.8\%), and Interruptibility (+20.7\%). 
Our contributions are summarized as follows:
\begin{itemize*}
\item \textbf{Cognitive Oversight}: We propose a human-AI collaboration paradigm: cognition oversight, which augments the intelligence through human-AI partnership. 
\item \textbf{Deep Cognition}: We operationalize the cognitive oversight paradigm into deep cognition, a multi-agent human-AI collaboration system designed for deep research tasks. 
\item \textbf{Comprehensive Evaluation Framework}: We establish a complete evaluation framework, including 15 metrics specifically designed for assessing the effectiveness of cognitive oversight in deep research scenarios. 
\item \textbf{Significant Performance Improvements}: Experiment results reveal significant improvements over strong deep research systems on both user evaluation and solving deep research problems. 
\end{itemize*}

\section{Related Work}


The relationship between humans and machines has undergone profound evolution over past decades. This section reviews human-AI interaction paradigms and recent deep research systems, highlighting differences with ours. 

\subsection{Human-AI Interaction}
We categorize human-AI interaction approaches as operational interaction~\cite{isinkaye2015recommendation, pazzani2007content, resnick1997recommender, fok2024search}, conversational interaction~\cite{openai2022chatgpt, mctear2016conversational, candello2016designing, openai2022chatgpt}, and mixed-initiative interaction~\cite{gervasio2025ai, bansal2024challengeshumanagentcommunication, bremers2024can, liu2025proactive, chen2025need}. Operational interaction emphasizes system-led automation with minimal user intervention, such as recommend systems and graphic interface~\cite{isinkaye2015recommendation, pazzani2007content, resnick1997recommender}. Conversational AI improves user alignment through contextual, multi-turn, and knowledge-driven dialogue—rather than merely executing predefined commands. Mixed-initiative interaction enables adaptive control over proactivity and reactivity through customizable interfaces, supporting dynamic coordination and user-system balance, especially in writing, coding and creativity tools~\cite{min2025feedforwardgenerativeaiopportunities}. Previous studies proposed interaction systems for mapping general-purpose AIs to the right specific use cases~\cite{jiang2024unknownunknownsengagedhuman, 10.1145/3544548.3581001,10.5555/3600270.3602446,jin2025disentanglingmemoryreasoningability, 10.1145/3706598.3714164, 10.1145/3706598.3713285, 10.1145/3706598.3713357, 10.1145/3708359.3712074, liu2025interactingthoughtfulai}. However, these current human-AI interaction approaches are still guided by the design principle of minimizing the need for human feedback, it shifts substantial cognitive burden to users. Humans still bear the cognitive burden of manual query formulation~\cite{10.1145/3654777.3676357, 10.1145/3654777.3676334}, result interpretation, knowledge synthesis~\cite{10.1145/3706598.3713953, 10.1145/3613904.3642698, Chen2024} and new idea generation~\cite{liu2024personaflowboostingresearchideation, 10.1145/3635636.3656184, osti_10523831, 10.1145/3613904.3642400} when they interact with these systems. 
Moreover, cognitive work was almost entirely human-driven, with systems serving as a mere instruction follower rather than an active human collaborator with system as an equality cooperative partner~\cite{shi2025pangudeepdiveradaptivesearch, zheng2025deepresearcher, jin2025searchr1, song2025r1searcher, chen2025ReSearch, XU2025103455, 10.1145/3706598.3713146, li2025confidencealignsexploringeffect}. As ~\citet{BAINBRIDGE1983775} warns, minimizing human feedback can erode operators’ skills and situational awareness—only to make their intervention when the system inevitably fails~\cite{yang2025riosworldbenchmarkingriskmultimodal, cemri2025multiagentllmsystemsfail}. 

As highlighted by ~\citet{white2024advancingsearchfrontierai} and ~\citet{feng2025levelsautonomyaiagents}, AI agents now support complex tasks through natural language interaction, better task understanding, and multi-level autonomy beyond basic queries interaction~\cite{srinivas2025scalingtesttimeinferencepolicyoptimized, shao2025futureworkaiagents}. The shift from static monolithic inference to adaptive, resource-aware computation has become central to AI systems for knowledge discovery~\cite{shao2024assistingwritingwikipedialikearticles, jiang2024unknownunknownsengagedhuman} leveraging multi-agent collaboration~\cite{watkins2025whatshumanhumanaicollaboration, fragiadakis2025evaluatinghumanaicollaborationreview} to facilitate serendipitous discovery. 
This mismatch constrains the potential for AI to act as a collaborator in exploratory inquiry~\cite{pirolli2009powers}. 
In this process, human experts readily adapt their investigative direction in response to unexpected findings as their understanding develops. Although current human-AI collaboration systems allow humans to read simplified model reasoning chains and engage in multi-turn asynchronous interactions with models~\cite{WESTPHAL2023107714, 10.3389/fcomp.2024.1521066, lee2024evaluatinghumanlanguagemodelinteraction, collins2024buildingmachineslearnthink}, these current interaction paradigms maintain limiting user's ability to adapt to emerging insights or evolving user understanding during complex and time-consuming tasks. 

\subsection{Deep Research Systems} 
\begin{table}[htbp]
\centering
{\small
\setlength{\tabcolsep}{2pt} 
\begin{tabular}{lccccccc}
\toprule
 & \textbf{Transparency} & \makecell{\textbf{Real-time}\\\textbf{Intervention}} & \makecell{\textbf{Fine-Grained}\\\textbf{Interaction}} & \makecell{\textbf{Preference}\\\textbf{Adaptation}} & \makecell{\textbf{Cognitive}\\\textbf{Oversight}}& \makecell{\textbf{Interactive}\\\textbf{Type}}  & \makecell{\textbf{Interaction-Driven}\\\textbf{Annotation}}\\
\midrule
\textbf{OpenAI} & \starfull \starhalf& \texttimes & \starfull & \texttimes & \starfull \starhalf& Input-Wait-Output & \texttimes \\
\textbf{Gemini} & \starfull \starfull& \texttimes & \starfull \starfull& \texttimes & \starfull \starfull& Input-Wait-Output & \texttimes \\
\textbf{Grok 3} & \starfull & \texttimes & \starfull & \texttimes & \starfull& Input-Wait-Output & \texttimes \\ 
\midrule
\textbf{DC.} & \starfull \starfull \starfull& \checkmark & \starfull \starfull \starfull& \checkmark & \starfull \starfull \starfull& Cognitive Interaction & \checkmark\\
\bottomrule
\end{tabular}
}
\caption{Comparison of different deep research systems (DC. indicates our deep cognition system).}
\label{tab:comparison_alt}
\end{table}

Deep research systems have garnered significant attention and adoption following the introduction of commercial platforms such as Gemini Deep Research~\cite{gemini_deep_research}, OpenAI Deep Research~\cite{openai_deep_research} and Grok3 Deeper Search~\cite{xai_grok_3}. These systems assist users in retrieving specific information for complex queries and conducting comprehensive, in-depth surveys on particular topics. Deep research capabilities are enabled by the sophisticated reasoning abilities that have emerged from recent advances in large language models (LLMs)~\cite{openai2024openaio1card, guo2025deepseek, kimiteam2025kimik15scalingreinforcement}, facilitating multi-step, in-depth analysis and information synthesis across hundreds of sources.
The implementation details of prominent commercial deep research systems remain proprietary and undisclosed. In contrast, most open-source deep research projects~\cite{open_deep_research, deep_research_dzhng, gpt_researcher, open_deep_research_camara, node_deepresearch, smolagents, deer_flow} employ prompt-based multi-agent systems with predefined workflows. Recent work~\cite{zheng2025deepresearcherscalingdeepresearch} has applied end-to-end reinforcement learning to open-source LLMs to perform iterative reasoning, web searching, and browsing, ultimately generating comprehensive answers to complex questions.
However, to the best of our knowledge, few existing deep research systems treat human-AI interaction during the research process as a first-class design consideration. While systems like Gemini Deep Research propose research plans or clarification questions to allow users to refine research objectives and requirements, user agency remains limited once research commences. Users cannot interrupt the ongoing process and can only engage through post-editing of generated reports or by asking follow-up questions after research completion. We summarize the comparisons with deep research systems in Table~\ref{tab:comparison_alt}.

The advancement toward increasingly sophisticated AI systems requires not merely enhanced model capabilities, but deeper integration between human and machine cognition. In this work, our proposed deep cognition system adopts a prompt-based multi-agent approach while incorporating extensive human interaction guided by the principles outlined in Section~\ref{sec:intro}.

\section{Deep Cognition}
We propose deep cognition, a multi-agent system designed to emulate human cognitive processes in deep research scenarios. This section provides a detailed description of our framework and its human-AI interaction mechanisms.

\begin{figure}[htbp]
    \centering
    \includegraphics[width=1\linewidth]{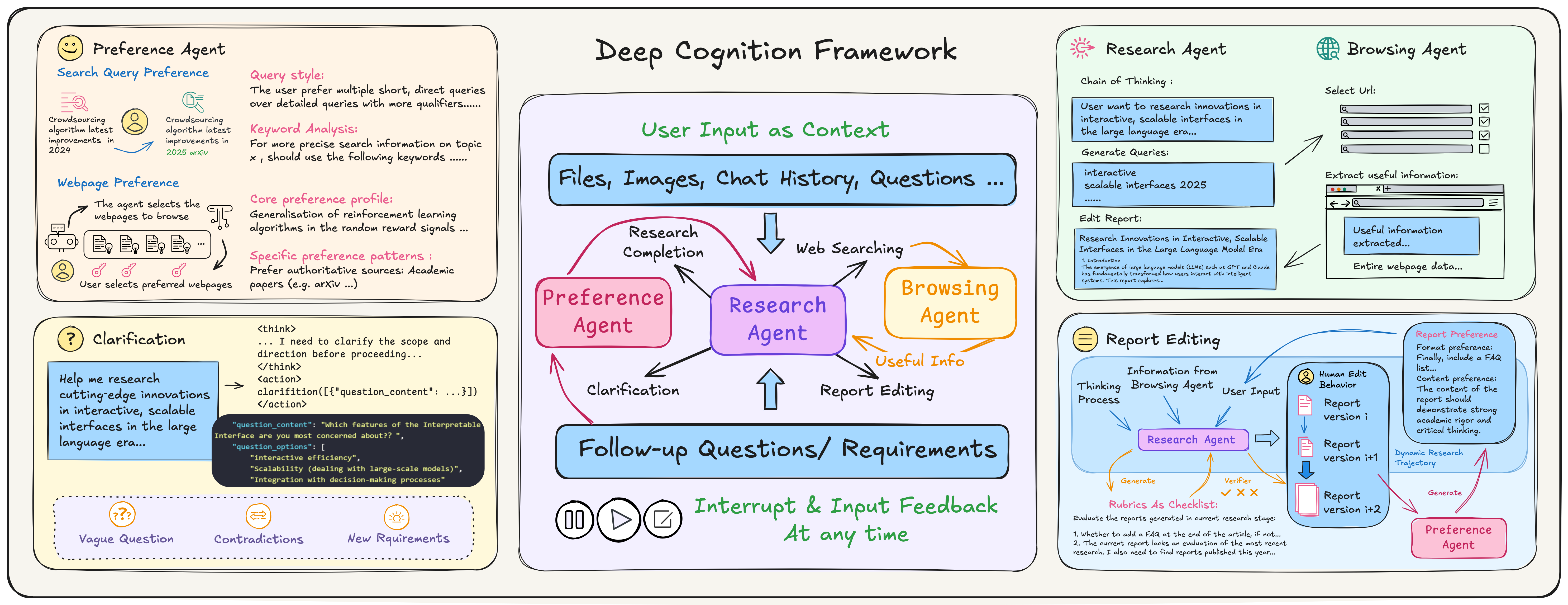}
    \caption{Deep cognition framework overview. This human-in-the-loop research assistant system breaks down complex research questions and dynamically synthesizes information from multiple sources through iterative search, clarification, and user feedback. The central \includegraphics[height=0.9em]{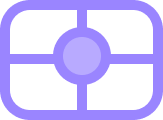} diagram illustrates the overall agent framework architecture. The framework integrates four key processes: \includegraphics[height=0.9em]{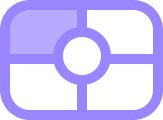} user preference modeling based on historical interaction patterns, \includegraphics[height=0.9em]{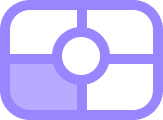} query refinement and clarification mechanisms, \includegraphics[height=0.9em]{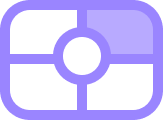} coordinated research orchestration, and \includegraphics[height=0.9em]{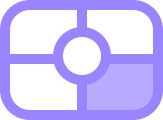} external environment interaction with automated report generation.}
    \label{fig:deep-cognition-framework}
\end{figure}

\subsection{The Deep Cognition Framework}

As illustrated in Figure~\ref{fig:deep-cognition-framework}, deep cognition is implemented as a multi-agent system comprising a \textbf{research agent}, a \textbf{browsing agent}, and a \textbf{preference agent}, supported by LLMs and search engine APIs. The research agent serves as the primary cognitive entity that collaborates with users to conduct deep research tasks, the browsing agent performs web browsing and gathers relevant information from multiple web pages with low latency and high accuracy, and the preference agent analyzes user trajectory to continuously adapt the system's behavior to user preferences.

\subsubsection{Research Agent} 
The research agent receives input context from users, including research questions, uploaded files (e.g., PDF files), images, or historical chat conversations. It conducts research through an interleaved thinking and action approach, leveraging large language models with advanced reasoning capabilities such as o1~\cite{openai2024openaio1card}, DeepSeek-R1~\cite{guo2025deepseek}, or Claude Sonnet 4~\cite{anthropic2025claudesonnet4}, etc. The research agent can execute the following actions during the research process:

\begin{itemize}
    \item \textbf{Clarification:} When the research agent encounters ambiguities in research questions, user requirements, or controversial content, it generates clarification questions for the user. We implement a presentation approach that displays notifications with multiple-choice options to reduce input burden while allowing users to provide detailed feedback when needed.

    \item \textbf{Web Searching:} Based on its own research trajectory, user requirements, preferences and the status of the current version of the report, it generates a list of targeted search queries and invokes the search engine API~\footnote{We use Google search in our implementation.} to retrieve the top $k$~\footnote{We set $k=5$ in this work.} for each query in parallel. Each search result consists of a title, snippet, and URL. These results are then processed by the Browsing Agent to extract relevant information.

    \item \textbf{Report Editing:} Once the agent gathers sufficient information from the browsing agent or receives editing requirements from the user, it generates a new version of the report by editing the current version while considering user preferences and requirements. To ensure the report content aligns with user preferences and requirements, the research agent proactively generates a list of quality rubrics for report verification beforehand. After generating the report, it systematically verifies each rubric and sends critiques back to itself for further refinement and polishing.

    \item \textbf{Research Completion:} When the agent determines that the report has covered sufficient aspects of the research question with adequate depth and has fulfilled the user's requirements, it concludes the research process. This decision is based on comprehensive assessment of content coverage, research depth, and alignment with stated objectives.
    
\end{itemize}

Through this iterative decision-making process, the research agent orchestrates the entire research workflow while maintaining focus on user needs and research quality. To support its information gathering needs, the system relies on the specialized capabilities of the browsing agent.

\subsubsection{Browsing Agent}
The browsing agent serves as a specialized information retrieval component that efficiently processes web search results and extracts relevant content to support the research agent's knowledge synthesis activities. This agent operates with high efficiency and accuracy to minimize latency while maximizing the quality of retrieved information. The browsing agent performs two primary functions during the research workflow:

\begin{itemize}
\item \textbf{URL Selection and Content Retrieval:} The agent employs intelligent filtering mechanisms to select URLs that demonstrate high potential for containing relevant information. This selection process analyzes multiple signals including the title, snippet, and URL structure of search results, while also incorporating user preferences regarding specific domains, topics, and source types. Once promising URLs are identified, the agent performs parallel web scraping operations to retrieve content efficiently. During this process, the system validates URL accessibility and maintains only those sources that can be successfully accessed, ensuring robust information gathering despite potential connectivity issues or restricted content.

\item \textbf{Information Extraction and Quality Assessment:} For each successfully retrieved webpage, the agent leverages large language models to perform sophisticated information extraction. This process involves identifying and extracting content segments that are directly relevant to the research objectives, while simultaneously generating concise summaries that capture the essence of useful information. Critically, the agent implements quality assessment mechanisms to distinguish between valuable and non-useful web pages, filtering out irrelevant content, advertisements, or low-quality sources that could diminish research quality.
\end{itemize}

The extracted information, accompanied by comprehensive metadata including titles and URLs for proper citation formatting, is transmitted back to the research agent. This structured approach ensures that the research agent receives high-quality, relevant information that can be seamlessly integrated into the report generation process while maintaining academic integrity through proper source attribution.

\subsubsection{Preference Agent}
The preference agent adapts to user preferences through the In-Context Reinforcement Learning paradigm~\cite{laskin2022incontextreinforcementlearningalgorithm, lin2024transformersdecisionmakersprovable, lee2023supervisedpretraininglearnincontext, huang2024incontextdecisiontransformerreinforcement, grigsby2024amagoscalableincontextreinforcement}, treating user actions and feedback as reward signals or critiques across three key dimensions:

\begin{itemize}
    \item \textbf{Query Preference:} The agent tracks user query modifications, refinements, and explicit search requirements to understand preferred search methodologies and terminology choices, subsequently influencing the research agent's query generation strategies.
    
    \item \textbf{Webpage Preference:} The agent monitors user selection patterns across different information sources and webpage types, identifying preferences for specific domains and publication types (e.g., academic papers vs. blogs). These learned patterns directly influence the browsing agent's decisions on URL selection.
    
    \item \textbf{Report Preference:} The agent analyzes user feedback, editing histories, and formatting requirements to understand preferences for report organization, writing style, and presentation approaches, guiding the research agent's report generation process.
\end{itemize}
    
This in-context adaptation mechanism enables the system to provide increasingly personalized research assistance by treating user interactions as reward signals, creating a dynamic research environment that evolves with user needs within each session.

\subsection{Interaction Design for Human-AI Collaboration}
\begin{figure}[htbp]
    \includegraphics[width=1\linewidth]{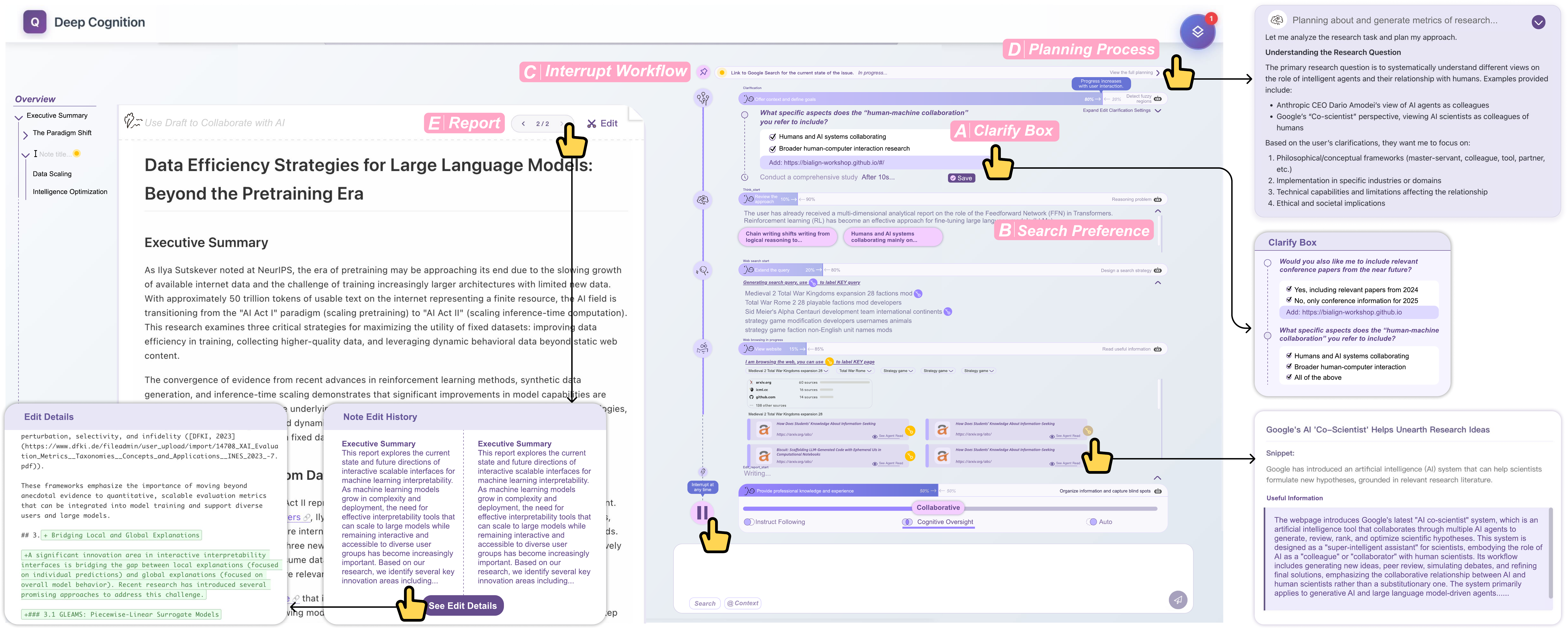}
    \caption{Deep cognition interface design showcasing key interactive features: (A) Research scope clarification to refine vague queries, (B) Preference searching based on user operation history and preferences, (C) Real-time human intervention capability, (D) Transparent display of reasoning, research processes, and interactive query refinement, and (E) Report revision area. The \includegraphics[height=1em]{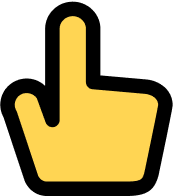} icon stands for clickable interface elements.}
    \label{fig:interaction-design}
\end{figure}
Deep research tasks are prevalent across domains such as investigative journalism, scientific research, and market analysis. These tasks are characterized by: (1) The necessity to synthesize information from multiple sources to address complex information-seeking problems or cover various facets of a topic; (2) Ongoing user interaction rather than single-query processing; (3) The production of curated, report-like artifacts rather than brief answers; and (4) Evolving user objectives that change iteratively as new information is accessed and integrated. Following principles of cognitive oversight, we designed the following features for our deep cognition system, with interfaces presented in Figure~\ref{fig:interaction-design}.

\paragraph{Feature I: Transparent Research Process}

The interface establishes transparency through multiple mechanisms that make the system's decision-making process visible and comprehensible to users. Search strategy explainability is achieved by directly displaying the reasoning process and query terms generated by the model, making information retrieval interpretable. This transparency enables users to understand how the system makes decisions and reaches conclusions. The editor area on the left of Figure~\ref{fig:interaction-design} displays the evolving research document with proper formatting. All findings are properly linked to their original sources, enabling users to trace source materials.

\paragraph{Feature II: Real-Time Intervention}

We implement a ``Pause'' feature that maintains user control throughout the research process, allowing users to interrupt the system at any point during execution. At critical junctures in the research process, users can provide feedback, introduce new requirements, or offer guidance to redirect the system's approach. This intervention capability enables users to actively shape the research trajectory based on emerging insights or changing objectives, ensuring the system remains aligned with user intentions throughout the investigation.

\paragraph{Feature III: Fine-Grained Interaction}

The interface supports multiple levels of user interaction to accommodate varying research needs and preferences. The system includes an ``Edit" button that enables direct modifications, allowing users to refine and restructure the document collaboratively. When users pose vague or unclear research questions, a clarification dialog appears with targeted questions to help narrow the scope---functioning like a research librarian asking ``What exactly are you looking for?". Users can refine their preferences and input follow-up knowledge or additional requirements as their understanding develops through the model's guidance. Users can prioritize preferred search queries and specify particular webpages or knowledge sources for emphasis. The search results visualization presents diverse results with source indicators, thumbnails, and organizational categorization to aid navigation.

\paragraph{Feature IV: User Research Preference Adaptation}
The system adapts to user preferences and requirements across multiple dimensions to provide personalized research assistance. Users can customize their preferences for report writing style, format, and structure. The system generates reference profiles based on user history and choices, learning from past interactions to better align with individual research approaches. Additionally, the system adapts to user preferences for search queries and knowledge sources, learning from selections and prioritizations to improve future recommendations. As research objectives evolve, the system adapts to new requirements while maintaining context from previous interactions. This adaptive capability ensures research process coherence while accommodating the natural evolution of deep research tasks.

Together, these four features create a comprehensive framework that balances automation with human oversight, enabling users to conduct thorough, efficient research while maintaining control over the process. By combining transparency, real-time intervention, fine-grained interaction, and adaptive personalization, the system empowers users to tackle complex research challenges with confidence and precision.

\section{Experiments}

We conducted user and benchmark evaluations to validate the effectiveness of our deep cognition framework's design principles. Our evaluation compared deep cognition against three leading deep research systems: Gemini Deep Research, OpenAI Deep Research, and Grok 3 Deeper Search. The assessment employed both user evaluation studies and quantitative analysis across two benchmark datasets. This section provides detailed descriptions of the evaluation methodologies, benchmarks, and metrics used in our analysis.

\subsection{User Evaluation}

We performed a user evaluation to capture real-world user experience during human-AI interaction inspired by \citet{lee2024evaluatinghumanlanguagemodelinteraction}. This methodology addresses two fundamental limitations of static benchmarks: 1) it reflects real-world, first-person subjective experience during human-AI interaction; and 2) it enables assessment of output quality that depends on interactive dynamics, which aligns with real-world usage scenarios.


\subsubsection{Protocol}

We hired 13 graduate students as participants for user evaluation. They have 30 minutes of time budget to interact with deep cognition to solve a specific problem they interested in. They also ask the other three systems the same question to get the corresponding answers. This user evaluation includes pre-study, in-study, and post-study.  The protocol consisted of three distinct phases: 
\paragraph{Pre-Study} The participants were introduced to the usage of deep cognition which familiarized participants with the system's interaction method. The detailed and complete user study protocol is provided in Appendix~\ref{sec:user-study-protocol}.

\paragraph{In-Study} When interacted with deep cognition, the participants actively reviewed both the model's reasoning processes and the evolving report draft. The session ended when deep cognition autonomously determined that the research was complete.

\paragraph{Post-Study} The participants completed a structured interview exploring their collaborative strategies. The interview examined three key dimensions, each includes a quantitative 5-point scales and several qualitative follow-up questions. 

\subsubsection{Evaluation Metrics Design}


The evaluation metrics we develop are shown in Table~\ref{tab:evaluation_metrics} which are inspired by OpenScholar~\cite{asai2024openscholarsynthesizingscientificliterature}. In addition, we define five key aspects in table~\ref{tab:evaluation_metrics} to evaluate the generated report. Each metric is rigorously assessed using a 5-point Likert scale. We assess organization, coverage, depth, relevance, helpfulness and cutting-edge, detailed definitions for each score (1–5) on the 5-point scale are provided in Appendix~\ref{sec:user-study-protocol}.

\begin{tcolorbox}[
    colback=white,
    colframe=custompurple,
    boxrule=1pt,
    sidebyside,
    sidebyside gap=8mm,
    sidebyside align=top,
    valign=center,
]

\scriptsize 
\renewcommand{\arraystretch}{1.3}
\begin{tabular}{>{\centering\arraybackslash}m{1cm}>{\RaggedRight\arraybackslash}m{5cm}}
\textbf{Metric} & \multicolumn{1}{c}{\textbf{Description}} \\
\midrule
Organization & Evaluate whether the article demonstrates sound organization and logical structure. An acceptable response should: 

(1) Exhibit clear structure by organizing relevant points into a coherent logical sequence. 

(2) Maintain coherence without any contradictions or unnecessary repetition. \\
\midrule
Cutting-Edge & Assess whether the article demonstrates comprehensive coverage of existing literature by: 

(1) Effectively summarizing and conducting comparative analysis with previous research.

(2) Timely incorporating the most recent and up-to-date research findings or information. \\
\midrule
Coverage & Provide comprehensive coverage of the identified areas of interest through: 

(1) Conducting thorough reviews.

(2) Citing a broad range of representative scholarly works.

(3) Incorporating the most current and time-sensitive information from various sources, rather than limiting the analysis to a small number of papers. \\
\midrule
Depth & Assess the adequacy of information content provided in the article. Specifically, evaluate whether the article delivers sufficient relevant information with appropriate depth such that readers can achieve thorough understanding of each argument presented. \\
\midrule
Relevance & Assess whether the response maintains topical relevance and preserves clear focus in order to deliver a useful response to the posed question. Specifically, the output should: 

(1) Sufficiently address the central elements of the original question and satisfy your informational requirements. 

(2) The response should exclude substantial amounts of tangential information unrelated to the original inquiry. \\
\end{tabular}

\tcblower

\scriptsize
\renewcommand{\arraystretch}{1.3}
\begin{tabular}{>{\centering\arraybackslash}m{1.5cm}>{\RaggedRight\arraybackslash}m{4.5cm}}
\textbf{Metric} & \multicolumn{1}{c}{\textbf{Description}} \\
\midrule
Intention to Use & Measures user intention and propensity for continued engagement with the system based on perceived value and satisfaction. \\
\midrule
Usability & Evaluates the intuitive nature and accessibility of the system interface, including cognitive load and interaction efficiency. \\
\midrule
Transparency & Assesses the interpretability and explainability of the model's decision-making processes and reasoning mechanisms. \\
\midrule
Interruptibility & Assesses the system’s ability to tolerate pauses or context switches and to resume smoothly without loss of state or progress. \\
\midrule
Fine-Grained Interaction & Evaluates the system's capacity to incorporate user feedback and enable precise, granular control over output generation. \\
\midrule
Inspiration & Assesses the system's ability to stimulate creative thinking and generate ideas or innovative approaches to problem-solving. \\
\midrule
Ease of Collaboration & Measures the extent to which the system functions as an effective collaborative partner in knowledge work and decision-making processes. \\
\midrule
Results-Worth-Effort & Evaluates whether users perceive the time and effort invested in system interaction as worthwhile and valuable relative to the outcomes achieved. \\
\midrule
Real-Time Intervention & Measures the degree to which users can actively interrupt and steer the system’s ongoing processes—e.g., pausing, editing, or re-prompting—to obtain desired outputs. \\
\midrule
Helpfulness & Assesses the overall utility and practical value of the output in addressing user needs and facilitating problem-solving objectives. \\
\end{tabular}
\end{tcolorbox}
\captionof{table}{Evaluation Metrics for Report Quality Assessment}
\label{tab:evaluation_metrics}

\subsection{Benchmark Evaluation}

To validate our hypothesis that experts with higher cognitive capabilities demonstrate enhanced collaboration with AI in transparent dialogue environments, we measured our system performance through the accuracy of browsecomp-ZH~\cite{zhou2025browsecompzhbenchmarkingwebbrowsing}, a benchmark assessing agents' web browsing capabilities. Given that our expert annotators are native Chinese speakers with domain expertise, we selected 22 questions (top two from each of 11 categories) for comprehensive assessment.

\section{Results and Analysis}

This section we introduce our evaluation experiments' results, including metrics design, user study design, user evaluation analysis and design suggestions. Our comprehensive evaluation demonstrates that deep cognition significantly outperforms existing systems across multiple dimensions. 

\subsection{Observations from User Evaluation}

\vspace{0.5em}
\begin{tcolorbox}[
    colback=custompurple!10,             
    colframe=custompurple,       
    boxrule=0.5pt,               
    arc=1.5mm,                     
    enhanced,
    shadow={1mm}{-1mm}{0mm}{shadowpurple}, 
    left=1em, right=1em, top=0.8em, bottom=0.8em
]

\textbf{\textit{Takeaway 1:}} By integrating transparency and interruptibility mechanisms at fine-grained interaction point of the research process, our system enables user intervention that measurably improves response quality—particularly in terms of \textbf{organization, cutting-edge, and depth}.
\end{tcolorbox}
\vspace{0.5em}

\noindent
\begin{minipage}[t]{0.45\textwidth}
As shown in Table~\ref{tab:deep_cognition_improvement}, augmented through expert interaction, the deep cognition system demonstrated significant enhancements across six evaluated metrics, overall average improve 63\%. Notably, the \textsc{organization} exhibits the greatest gain (+97\%), followed by \textsc{cutting-edge} (+79\%) and depth (+76\%). Even the dimension with the smallest gain, helpfulness, showed a significant improvement of +42\%. As the evaluation results in Table~\ref{tab:report evaluation}, the \textbf{alignment between expert rankings and user evaluations} validates our core hypothesis: \textbf{The system with enhanced interaction mechanisms consistently deliver output quality across six metrics.}

\end{minipage}
\hfill
\begin{minipage}[t]{0.5\textwidth}
\vspace{-0.5em}
\centering
\small
\begin{tabular}{lcc}
\toprule
\multicolumn{1}{c}{\textbf{Metric}} & \textbf{DC (non).} & \textbf{DC.} \\
\midrule
Organization  & \textbf{2.231}    & \textbf{4.385} \textcolor{lightorange}{\textbf{↑ 97\%}}\\
Cutting-Edge  & \textbf{2.538}    & \textbf{4.538} \textcolor{lightorange}{\textbf{↑ 79\%}}\\
Coverage      & 2.423    & 4.000 \textcolor{lightorange}{↑ 65\%}\\
Depth         & \textbf{2.231}    & \textbf{3.923} \textcolor{lightorange}{\textbf{↑ 76\%}}\\
Relevance     & 2.885    & 3.769 \textcolor{lightorange}{↑ 31\%}\\
Helpfulness   & 2.808    & 4.000 \textcolor{lightorange}{↑ 42\%}\\
\midrule
Overall Average & 2.519  & 4.103 \textcolor{lightorange}{↑ 63\%}\\
\bottomrule
\end{tabular}
\vspace{-0.5em}
\captionof{table}{\footnotesize Performance improvement of deep cognition over deep cognition without interaction. DC. indicates deep cognition, DC (non). indicates deep cognition without interaction.}
\label{tab:deep_cognition_improvement}
\end{minipage}
\vspace{0.5em}

\begin{table}[h]
\small
\renewcommand{\arraystretch}{1.3}
\setlength{\tabcolsep}{2pt}         
\begin{minipage}[t]{0.48\linewidth}
{\centering                    
  Report Evaluation (1–5 Score)\\[4pt]\par}
\raggedright  
\begin{tabular}{lcccc}
\toprule
\multicolumn{1}{c}{\textbf{Metric}} & \textbf{DC.} & \textbf{Gemini} & \textbf{OpenAI} & \textbf{Grok3} \\
\midrule
Organization    
& \cellcolor{red2}\textbf{4.385}\textsubscript{\scriptsize\textbf{\textcolor{lightorange}{+1.8\%}}}
& \cellcolor{red1}4.308 
& \cellcolor{blue2}3.769 
& \cellcolor{blue1}3.385 \\
Cutting-Edge    
& \cellcolor{red2}\textbf{4.538}\textsubscript{\scriptsize\textbf{\textcolor{lightorange}{+3.5\%}}}
& \cellcolor{red1}4.385 
& \cellcolor{blue2}3.769 
& \cellcolor{blue1}3.538 \\
Coverage        
& \cellcolor{red1}4.000\textsubscript{\scriptsize\textbf{\textcolor{lightmint}{-10.4\%}}}
& \cellcolor{red2}\textbf{4.462} 
& \cellcolor{blue2}3.692 
& \cellcolor{blue1}2.923 \\
Depth           
& \cellcolor{red1}3.923\textsubscript{\scriptsize\textbf{\textcolor{lightmint}{-1.9\%}}}
& \cellcolor{red2}\textbf{4.000} 
& \cellcolor{blue2}3.577 
& \cellcolor{blue1}2.769 \\
Relevance       
& \cellcolor{blue2}3.769\textsubscript{\scriptsize\textbf{\textcolor{lightmint}{-18.3\%}}}
& \cellcolor{red2}\textbf{4.615} 
& \cellcolor{red1}4.077 
& \cellcolor{blue1}3.615 \\
Helpfulness     
& \cellcolor{red2}\textbf{4.000}\textsubscript{\scriptsize\textbf{\textcolor{lightorange}{+0.0\%}}}
& \cellcolor{red2}\textbf{4.000} 
& \cellcolor{red1}3.615 
& \cellcolor{blue1}2.692 \\
\bottomrule
\end{tabular}
\end{minipage}
\hspace{1pt} 
\begin{minipage}[t]{0.48\linewidth} 
\centering
Interaction Evaluation (1–5 Score)\\[4pt]
\begin{tabular}{lcccc}       
\toprule
\multicolumn{1}{c}{\textbf{Metric}} & \textbf{DC.} & \textbf{Gemini} & \textbf{OpenAI} & \textbf{Grok\,3} \\
\midrule
Transparency
& \cellcolor{red2}\textbf{\underline{5.00}}\textsubscript{\scriptsize\textbf{\textcolor{lightorange}{+25.0\%}}}
& \cellcolor{red1} 4.00
& \cellcolor{blue1} 3.00
& \cellcolor{blue2} 3.19 \\

Interruptibility
& \cellcolor{red2}\textbf{4.35}\textsubscript{\scriptsize\textbf{\textcolor{lightorange}{+31.4\%}}}
& \cellcolor{red1} 3.31
& \cellcolor{blue2} 2.69
& \cellcolor{blue1} 2.62 \\

Fine-Grained Interation
& \cellcolor{red2}\textbf{\underline{4.73}}\textsubscript{\scriptsize\textbf{\textcolor{lightorange}{+44.6\%}}}
& \cellcolor{red1} 3.27
& \cellcolor{blue2} 2.88
& \cellcolor{blue1} 2.19 \\

Real-Time Intervention
& \cellcolor{red2}\textbf{\underline{4.69}}\textsubscript{\scriptsize\textbf{\textcolor{lightorange}{+24.4\%}}}
& \cellcolor{red1} 3.77
& \cellcolor{blue2} 2.92
& \cellcolor{blue1} 2.62 \\

Inspiration
& \cellcolor{red2}\textbf{4.08}\textsubscript{\scriptsize\textbf{\textcolor{lightorange}{+0.0\%}}}
& \cellcolor{red2}\textbf{4.08}
& \cellcolor{red1} 3.42
& \cellcolor{blue2} 3.19 \\

Ease of Collaboration
& \cellcolor{red2}\textbf{4.62}\textsubscript{\scriptsize\textbf{\textcolor{lightorange}{+43.0\%}}}
& \cellcolor{red1} 3.23
& \cellcolor{blue2} 2.77
& \cellcolor{blue1} 1.85 \\


Results-Worth-Effort
& \cellcolor{red2}\textbf{4.52}\textsubscript{\scriptsize\textbf{\textcolor{lightorange}{+10.8\%}}}
& \cellcolor{red1} 4.08
& \cellcolor{blue2} 3.29
& \cellcolor{blue1} 2.96 \\
\bottomrule
\end{tabular}
\label{tab:system_scores_inline}
\end{minipage}

\caption{User and expert evaluation results for AI research assistance systems. Left panel: User-generated evaluation scores on a 1-5 scale, where participants queried systems with their own research questions (n=13 participants, 13 responses). Right panel: Scores (1–5 scale) for system-interaction evaluation metrics (n = 13 participants). Color coding indicates within-row performance rankings, and percentages show deep cognition’s relative improvement over the strongest baseline system (Gemini). DC. indicates deep cognition.}
\label{tab:report evaluation}
\end{table}
\vspace{0.8em}  
\noindent


Deep cognition dominates six of the seven metrics. It records the largest gains in Fine-Grained Interaction (+44.6\%) and Cooperative (+43.0\%), and is the only system to reach a perfect Transparency score (5.00, +25.0\% over the strongest baseline). Overall, the results highlight deep cognition’s superior transparency, controllability, and collaborative support. These quantitative results are further supported by users' qualitative feedback. Over 90\% of participants agree or strongly agree that interaction with deep cognition improves report quality; 69\% find it easy to use and 62\% show a high willingness to use. 

\begin{figure}[!h]
    \centering
    \includegraphics[width=1\linewidth]{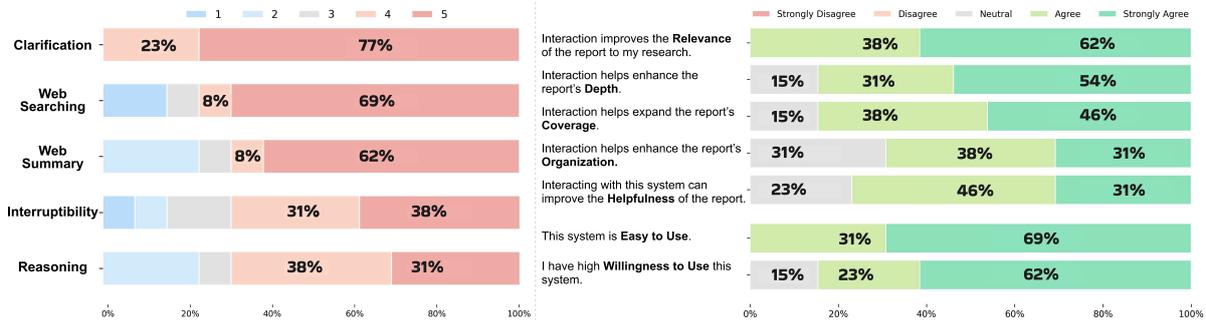}
    \caption{Left: Distribution of participant ratings (1–5) indicating the extent to which each system feature benefited their research process (n = 13 participants). Right: Perceived overall usefulness of deep cognition, as reported by the same participant cohort (n = 13 participants).}
    \label{fig:user_evaluation}
\end{figure}

\begin{tcolorbox}[
    colback=lightorange!10,             
    colframe=orange,       
    boxrule=0.5pt,               
    arc=1.5mm,                     
    enhanced,
    shadow={1mm}{-1mm}{0mm}{deeporange}, 
    left=1em, right=1em, top=0.8em, bottom=0.8em
]

\textbf{\textit{Design Suggestion 1:}} Usage as Annotation becomes possible through thoughtful product design that transforms natural user interactions into annotation signals. When users complete tasks, their behaviors implicitly provide annotation signals that guide system adaptation. Complex user hesitations or corrections trigger deeper reasoning processes, while smooth task completion indicates successful lightweight inference. 

\end{tcolorbox}

\subsection{Observation from Benchmark Evaluation}

\begin{tcolorbox}[
    colback=custompurple!10,             
    colframe=custompurple,       
    boxrule=0.5pt,               
    arc=1.5mm,                     
    enhanced,
    shadow={1mm}{-1mm}{0mm}{shadowpurple}, 
    left=1em, right=1em, top=0.8em, bottom=0.8em
]

\textbf{\textit{Takeaway 2:}} Participants with deeper cognitive processing capabilities achieved significantly higher human-AI collaborative performance compared to those with surface-level cognitive approaches in transparent interaction paradigms, as measured by problem resolution accuracy.. 

\end{tcolorbox}

\begin{table}[h]
  \centering
  \begin{threeparttable}
    \begin{tabular}{lcccccc}
      \toprule
                    & \textbf{DC (non cog).} & \textbf{DC (non int).} &
                      \textbf{DC (cog+int).} & \textbf{Gemini} & \textbf{OpenAI} & \textbf{Grok 3} \\
      \midrule
      Accuracy   & 45.45\% & 40.91\% & 72.73\% & 40.91\% & 40.91\% & 22.73\% \\
      \bottomrule
    \end{tabular}
  \end{threeparttable}
    \caption{Accuracy comparison across deep research systems. DC (non cog). represents the baseline condition with participants possessing foundational knowledge levels (n=4 participants with middle school-level knowledge); DC (non int). represents the autonomous system condition without human intervention; DC (cog+int). represents the interactive condition with graduate-level participants engaging in real-time collaboration with the system (n=4 participants). }
\label{benchmark_evaluation}
\end{table}

The results in Table~\ref{benchmark_evaluation} provide compelling evidence for our collaborative cognition framework. The deep cognition system (cognition + interaction) achieves 72.73\% accuracy —— a dramatic improvement over all baselines and ablated versions. This performance significantly exceeds standalone systems: Gemini and OpenAI both plateau at 40.91\%, while Grok 3 lags at 22.73\%. Critically, the ablation study reveals that neither cognitive oversight alone (45.45\%) nor interaction capabilities alone (40.91\%) approach the performance of their combination, demonstrating that expert-AI collaboration requires both transparent reasoning processes and fine-grained interactive guidance to tackle challenging web browsing tasks effectively.

\subsection{Dynamic Autonomy in Human-AI Deep Research System}
\begin{figure}[htbp]
    \centering
    \includegraphics[width=1\linewidth]{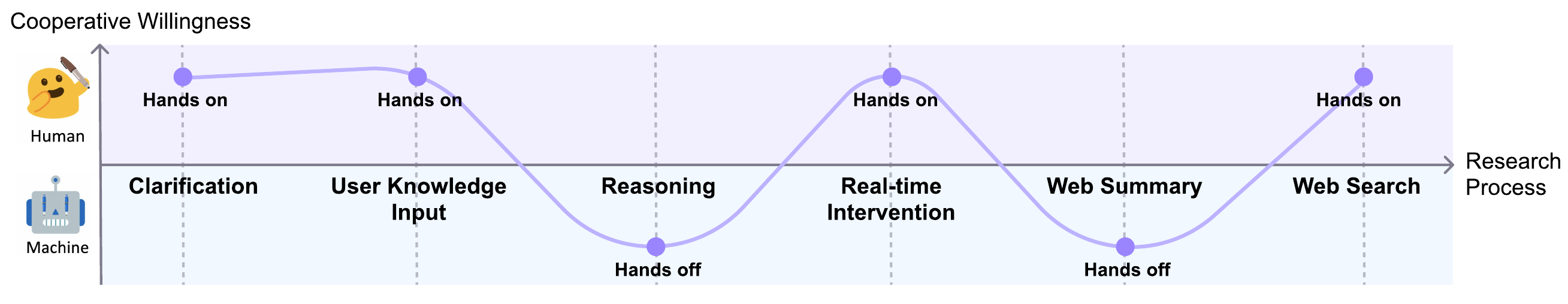}
    \caption{Changes in users’ behavioral tendencies when using the deep research system to perform complex retrieval tasks.}
    \label{fig:behavior}
\end{figure}

We dive deeper into the human behavior pattern in the deep research process and provide design considerations of human-AI collaboration research system. As illustrated in Figure~\ref{fig:behavior}, our user study reveals a sophisticated pattern of collaborative engagement that varies systematically across six research phases. Users demonstrate \textbf{dynamic cooperation willingness}, transitioning between ``hands-on" and ``hands-off" modes based on task characteristics and their domain expertise. We detail these six phases below: 
\begin{tcolorbox}[
    colback=lightorange!10,             
    colframe=orange,       
    boxrule=1pt,               
    arc=1.5mm,                     
    enhanced,
    shadow={1mm}{-1mm}{0mm}{deeporange}, 
    left=1em, right=1em, top=0.8em, bottom=0.8em
]

\textbf{\textit{Design Suggestion 2:}} Enhancing transparency at the model's behavioral status can improve human-AI collaboration. Specifically, in complex, long-duration retrieval tasks, humans tend to delegate mechanical operations such as ``browsing'' and ``summarizing'' to AI, while preferring to collaborate with the model at decision points requiring higher-order thinking.

\end{tcolorbox}

\paragraph{Phase I: Clarification (Hands-on)} The research process begins with intensive human-AI collaboration as users refine vague problem definitions. Users' initial research questions are typically too broad to cover all possible scenarios. For example, participant P1's first query was deliberately open-ended: \textit{``What are the current mainstream approaches to repo-level code generation and completion?"} which they then refined with \textit{``What about recent popular techniques such as Claude Code, Windsurf, and Cursor?"} This iterative refinement process proved valuable---8 participants mentioned that the model filled in aspects beyond their expectation. As P1 reflected: \textit{``I aligned my thinking purpose with others. The question did contain aspects that I had not thought of before"} during the clarification stage.

\paragraph{Phase II: User Knowledge Input (Hands-on)} Users maintain high engagement when they possess specific domain knowledge or references that need integration. When users know specific references or attributes about an item, such as queries, paper links, websites, or personal opinions, they actively guide the AI to relevant media. This is exemplified by targeted directives such as \textit{``This website is of critical importance and should be reviewed thoroughly"}.

\paragraph{Phase III: Reasoning (Hands-off)} Interestingly, users demonstrate lower cooperative willingness during reasoning phases, preferring to assess the AI's autonomous analytical capabilities. Users seek to understand whether the model has correctly executed prescribed instructions and want transparency in decision-making processes. P12 emphasized wanting to see \textit{``how the model decides on key technical routes when faced with some open-ended problems"}. This hands-off approach allows users to evaluate AI reasoning quality without excessive intervention.

\begin{tcolorbox}[
    colback=lightorange!10,             
    colframe=orange,       
    boxrule=0.5pt,               
    arc=1.5mm,                     
    enhanced,
    shadow={1mm}{-1mm}{0mm}{deeporange},  
    left=1em, right=1em, top=0.8em, bottom=0.8em
]

\textbf{\textit{Design Suggestion 3:}} Optimal human-AI collaboration requires cognitively appropriate responses tailored to users' expertise levels, rather than merely preference alignment. Our findings show that challenging model outputs motivate users to contribute additional domain knowledge, enhancing collaborative outcomes.

\end{tcolorbox}

\paragraph{Phase IV: Real-Time Intervention (Hands-on)} Cooperation peaks again during dynamic browsing tasks where users encounter pages or information sources that warrant detailed retrieval. Users actively adjust the model's responses to align with their expectations.

\paragraph{Phase V: Web Summary (Hands-off)} During summarization tasks, users tend to trust in AI capability. Participants often need consolidated insights from multiple sources rather than single source summarization, leading them to allow extended autonomous operation. As P12 noted: \textit{``I will assume that AI can find useful information by itself as long as I look at it".} This reflects sophisticated understanding of when mechanical aggregation tasks can be delegated effectively.

\paragraph{Phase VI: Web Search (Hands-on)} The cycle concludes with renewed hands-on engagement for open-ended and subjective questions that require interpretation or subjective judgment. These prove most challenging for existing tools, where conventional approaches require ongoing human guidance throughout the process. However, cognitive oversight allows extended periods of autonomous AI cognition punctuated by targeted human interventions at critical junctures.

This dynamic pattern demonstrates that effective human-AI collaboration is not uniform but adapts strategically to leverage the comparative advantages of human judgment and AI processing capabilities across different research phases. We illustrate this dynamic research task example to demonstrate authentic participant behavior.

\begin{tcolorbox}[colback=custompurple!5, 
    colframe=custompurple, 
    boxrule=1pt,               
    arc=1.5mm,  
  coltitle=black!80, 
  title=\textbf{A Dynamic Research Task Case},
  colbacktitle=custompurple, fonttitle=\bfseries]
\small
\textbf{Domain}: Interdisciplinary Writing
\tcbline 
\textbf{Initial query}: Please assist in investigating the latest innovations in interactive and scalable interfaces designed to enhance the interpretability of large language models in writing.
\tcbline 
\textbf{Clarify query}: What aspect of interpretability are you most interested in exploring? Who is the primary target audience for these interfaces? What writing contexts are you most interested in? What aspects of interface scalability are most relevant to your research? Are you interested in any specific emerging technologies related to LLM interpretability?
\tcbline 
\textbf{Domain knowledge input}: Add Jeff Rzeszotarski's PhD dissertation, and research in PAIR (People + AI Research Initiative) team.
\tcbline 
\textbf{Initial goal}: Development trend of interpretability of Interpretable Machine Learning Interface
\tcbline
\textbf{Last goal}: Investigate which research fields the scholars who previously worked in this direction have migrated to.
\end{tcolorbox}

\section{Conclusion}

This paper introduced deep cognition, a human-AI collaboration system that implements ``cognitive oversight" through transparent, interruptible interactions. Rather than treating interaction as merely an interface, we demonstrated that interaction constitutes a fundamental dimension of intelligence for complex research tasks.

Our evaluation shows that cognitive oversight significantly outperforms traditional approaches, achieving 63\% average improvement over non-interactive systems and 72.73\% accuracy on BrowseComp-ZH benchmark. The system excels particularly in transparency (+25.0\%), fine-grained interaction (+44.6\%), and cooperative capabilities (+43.0\%). User behavior analysis revealed sophisticated dynamic autonomy patterns, where participants strategically alternate between ``hands-on" and ``hands-off" modes across different research phases.

These findings challenge the assumption that AI progress requires purely autonomous capabilities. Instead, our work suggests that advanced intelligence emerges from cognitive partnerships that leverage complementary human judgment and machine processing strengths. This establishes the foundation for reconceptualizing human-AI relationships in complex cognitive tasks.

\section*{Contribution}

\paragraph{Project Lead}
Lyumanshan Ye, Xiaojie Cai

\paragraph{Algorithm}
Xiangkun Hu, Xinkai Wang, Sihan Wang, Bohan Zhang, Xiaoze Fan, Jinbin Luo

\paragraph{UI Interface}
Junfei Wang, Yang Nan, Jiadi Su

\paragraph{Annotation}
Yuxiang Zheng, Tianze Xu, Dayuan Fu, Yunze Wu, Pengrui Lu, Zengzhi Wang, Yiwei Qin, Zhen Huang, Yan Ma, Zhulin Hu, Haoyang Zou, Tiantian Mi, Yixin Ye, Ethan Chern

\paragraph{Supervision \& Corresponding Author}
Pengfei Liu


\clearpage

\bibliographystyle{acl_natbib}
\bibliography{related}

\clearpage

\appendix

\section{User Behavior Data Point}
\begin{figure}[!ht]
  \centering
  \includegraphics[width=0.7\linewidth]{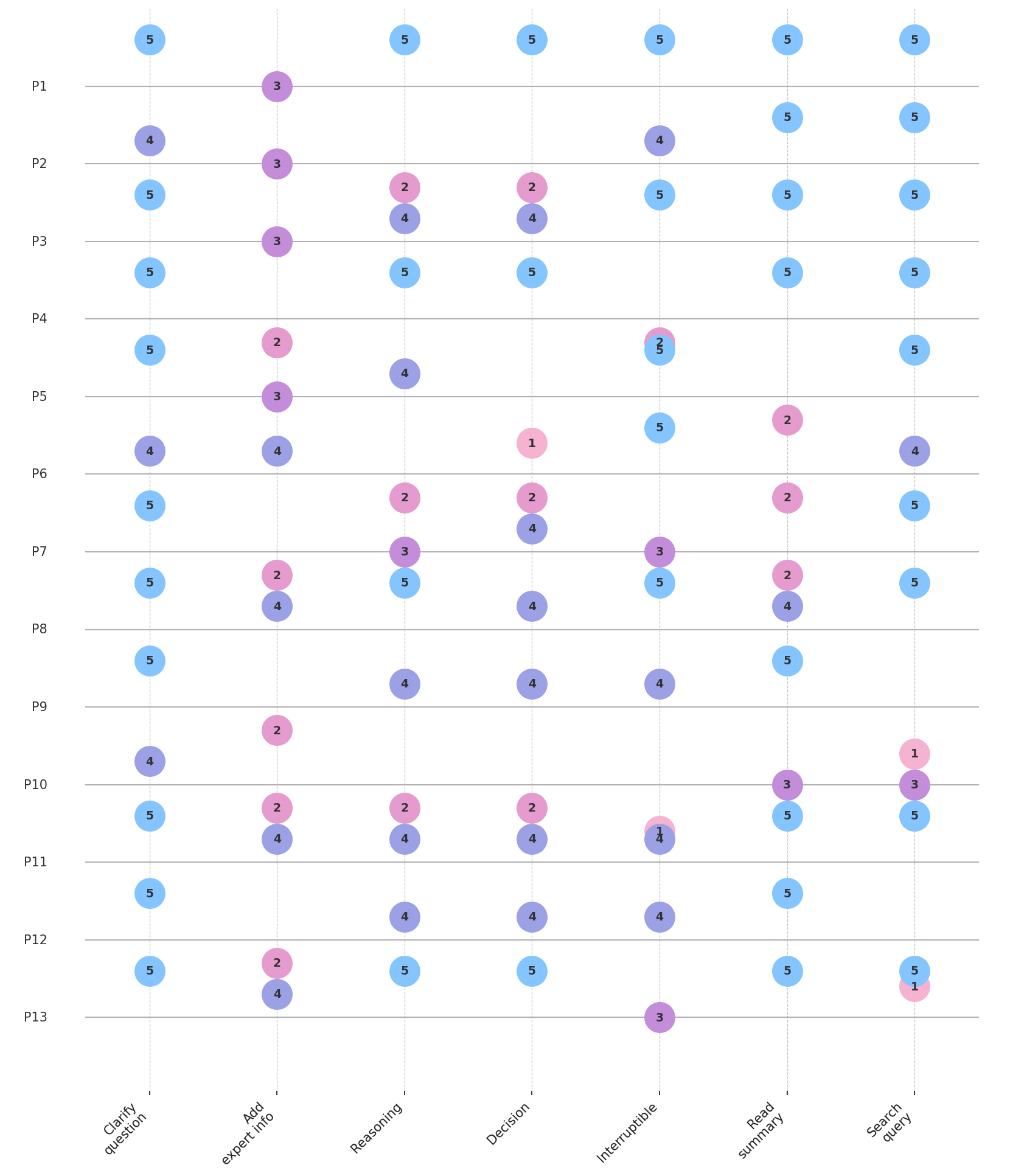}
  \caption{Human–AI collaboration code book}
  \label{fig:ai collaboration code book}
\end{figure}

\section{User Study Protocol}\label{sec:user-study-protocol}
\subsection{Pre-Study}

\paragraph{Study Overview}

This protocol evaluates four AI research systems: deep cognition, OpenAI Deep Research (O3), Grok 3 Deeper Search, and Gemini Deep Research (default). Participants complete authentic research tasks requiring between 15 and 30 minutes per system, with a maximum interaction time of 30 minutes allocated to deep cognition. 

\paragraph{Participant Instructions}

Thank you for helping us conduct this evaluation. You need to pose a research question that you genuinely want to ask. Typically, this research question should be somewhat ambiguously defined, focused on open-ended inquiry, with substantial room for interpretation in the response, and requiring iterative search and adjustment. For example:

\textit{``I want to systematically understand current perspectives on how to position 'AI agent roles and their relationships with humans.' For instance, Anthropic CEO Dario Amodei believes that future AI agents will relate to humans as colleagues; Google published a paper on Co-scientist, viewing AI scientists as human colleagues. Please collect more viewpoints and analyze them in combination with current and future development trends."}

\textit{``Why can models trained on synthetic data outperform models that provide synthetic data? Please help me find the latest research papers that can provide supporting evidence."}
Typically, a report may take 15-30 minutes to generate, with a maximum time limit of 30 minutes for Deep Cognition interaction. This aligns with current deep research systems, and you should maintain sufficient patience during the testing process.

\textit{``Ilya mentioned at NeurIPS that pretraining is approaching its end because internet data is not growing at a particularly fast rate, and models currently lack sufficient new data to satisfy the training of larger models. Therefore, a current challenge is how to improve data utilization efficiency (as mentioned by OpenAI researchers) - assuming there are approximately 50T tokens of data on the internet, how can we utilize these 50T tokens effectively to improve the intelligence ceiling of models? Please help me research relevant materials and literature, identifying methods for improving data utilization efficiency and ways to collect more data. For example, current web data is static - how might we obtain dynamic data, such as behavioral traces?"}

\paragraph{1. Pre-Study (Understanding System Usage)}
This is a tool for real-time human-AI collaboration, retrieving open-ended multi-hop questions, allowing users to dynamically explore initial questions during system interaction and ultimately complete comprehensive writing. Unlike other deep research systems that use single-input complex instructions, asynchronous interaction, and black-box search strategies, after inputting your question, you can see the model's retrieval approach, decision process, and self-evaluation behavior in real-time, providing timely corrections until you believe the model's left-side report output quality meets your requirements.

You cannot directly manually modify the model's final report. You need to guide the model to improve report writing depth and information retrieval efficiency through various interaction methods during the model's research process (interruption, adding expert prior knowledge, reviewing model-retrieved information, auditing the model's self-evaluation process, new thinking, strategic guidance, or personal files). Please note that you should aim to achieve 4-5 points across all dimensions before stopping generation. You can interrupt at any time before the model finishes. The termination point is when the model autonomously decides to finish.

\textit{Model Settings: After selecting ``Clarify Question" copy and record the thought chain returned on the right side.
You need to simultaneously review the behavioral patterns returned by the model on the right side.
When using Deep Cognition, you need to enable the switch in the bottom right corner.}

\subsection{In-Study}
\paragraph{Understanding Evaluation Metrics} During generation across all systems, you need to timely review the model's behavior (right-side thought chains, expanded model execution details, all searched URLs, information retrieved from URLs) and the quality of model-generated reports (left-side drafts).

\subsubsection{Evaluation Framework}

\begin{table}[h]
\centering
\renewcommand{\arraystretch}{1.5}
\begin{tabular}{p{8cm}*{5}{>{\centering\arraybackslash}p{1cm}}}
\textbf{Evaluation Dimension} & \textbf{Pool} & \textbf{Basic} & \textbf{Average} & \textbf{Strong} & \textbf{Exceptional} \\
\hline
\textbf{Organization:} Structural clarity and logical flow & $\bigcirc$ & $\bigcirc$ & $\bigcirc$ & $\bigcirc$ & $\bigcirc$ \\
\hline
\textbf{Cutting-edge Information:} Coverage of recent, high-impact research & $\bigcirc$ & $\bigcirc$ & $\bigcirc$ & $\bigcirc$ & $\bigcirc$ \\
\hline
\textbf{Information Coverage (Breadth):} Comprehensiveness across research domains & $\bigcirc$ & $\bigcirc$ & $\bigcirc$ & $\bigcirc$ & $\bigcirc$ \\
\hline
\textbf{Information Depth:} Sufficiency of detail for thorough understanding & $\bigcirc$ & $\bigcirc$ & $\bigcirc$ & $\bigcirc$ & $\bigcirc$ \\
\hline
\textbf{Overall Helpfulness:} Practical utility for literature review and research & $\bigcirc$ & $\bigcirc$ & $\bigcirc$ & $\bigcirc$ & $\bigcirc$ \\
\end{tabular}
\caption{5-Point Likert Scale for Assessing Report Quality}
\end{table}

\begin{tcolorbox}[title=Organization, colback=gray!10, colframe=customdeepPurple!50, rounded corners, coltitle=black,
fonttitle=\bfseries, boxrule=1pt, width=\textwidth, breakable]
\setlength{\parskip}{0pt}
\textbf{Definition}
    Evaluate whether the article has good organization and logical structure. An acceptable response should: 1. Have clear structure, categorizing related points into a logical flow. 2. Be coherent, without contradictions or unnecessary repetition.
\vspace{0.5em}

\textbf{Score 5: Exceptional Organization}
\begin{itemize}[topsep=0pt, partopsep=0pt, itemsep=0pt]
    \item \textbf{Structure Clarity:} Perfect logical structure with clear hierarchical organization and seamless section transitions;
    \item \textbf{Logical Flow:} Flawless reasoning progression from introduction to conclusion with excellent coherence;
    \item \textbf{Coherence:} All content elements perfectly interconnected with consistent thematic development;
    \item \textbf{Presentation Quality:} Outstanding formatting and layout that enhances readability and comprehension;
\end{itemize}
    \vspace{0.5em}
    \textbf{Score 4: Strong Organization}
    \begin{itemize}[topsep=0pt, partopsep=0pt, itemsep=0pt]
        \item \textbf{Structure Clarity:} Response is well-organized with clear, logical structure consistently followed;
        \item \textbf{Logical Flow:} Points are effectively grouped, flow is smooth;
        \item \textbf{Coherence:} Minor coherence issues but overall clear and easy to follow with minimal repetition or contradictions;
        \item \textbf{Presentation Quality:} Good formatting that supports understanding;
    \end{itemize}
    \vspace{0.5em}
    \textbf{Score 3: Moderate Organization}
    \begin{itemize}[topsep=0pt, partopsep=0pt, itemsep=0pt]
        \item \textbf{Structure Clarity:} Response is generally well-organized with clear structure that is basically maintained;
        \item \textbf{Logical Flow:} Adequate progression with some choppy transitions;
        \item \textbf{Coherence:} Reasonable thematic development with some disconnected elements;
        \item \textbf{Presentation Quality:} Acceptable formatting with room for improvement;
    \end{itemize}
    \vspace{0.5em}
    \textbf{Score 2: Basic Organization}
    \begin{itemize}[topsep=0pt, partopsep=0pt, itemsep=0pt]
        \item \textbf{Structure Clarity:} Some organization but inconsistent structure, minor contradictions;
        \item \textbf{Logical Flow:} Weak reasoning progression with confusing transitions;
        \item \textbf{Coherence:} Limited thematic coherence with noticeable gaps;
        \item \textbf{Presentation Quality:} Poor formatting that hinders comprehension;
    \end{itemize}
    \vspace{0.5em}
    \textbf{Score 1: Poor Organization}
    \begin{itemize}[topsep=0pt, partopsep=0pt, itemsep=0pt]
        \item \textbf{Structure Clarity:} No clear structure, scattered points, difficult to follow;
        \item \textbf{Logical Flow:} No discernible logical progression, chaotic presentation;
        \item \textbf{Coherence:} No thematic coherence, completely disconnected content;
        \item \textbf{Presentation Quality:} Very poor formatting that severely impairs understanding;
    \end{itemize}
\end{tcolorbox}

\begin{tcolorbox}[title=Cutting-Edge Information, colback=gray!10, colframe=customdeepPurple!50, rounded corners, coltitle=black,
fonttitle=\bfseries, boxrule=1pt, width=\textwidth, breakable]
    \setlength{\parskip}{0pt}

\textbf{Definition}
    Evaluate whether the article effectively summarizes the past, compares with previous research, and timely identifies the latest, most current research or information.
\vspace{0.5em}

    \textbf{Score 5: Exceptional}
    \begin{itemize}[topsep=0pt, partopsep=0pt, itemsep=0pt]
        \item \textbf{Recency:} Precisely captures key latest research in the field, including recently published technical reports, preprints, conference reports, and ongoing work;
        \item \textbf{Impact Level:} Includes highest-impact research and breakthrough discoveries, keen insight into cutting-edge issues and breakthrough progress, can identify emerging directions not yet widely recognized;
        \item \textbf{Coverage Completeness:} Comprehensive coverage of all major recent developments;
        \item \textbf{Source Quality:} Exclusively high-quality, authoritative sources from leading institutions;
    \end{itemize}
    \vspace{0.5em}
    \textbf{Score 4: Strong}
    \begin{itemize}[topsep=0pt, partopsep=0pt, itemsep=0pt]
        \item \textbf{Recency:} Response successfully identifies most important recent research achievements and breakthrough work;
        \item \textbf{Impact Level:} Covers major high-impact developments with good selection. Has clear grasp of recent developments, can precisely identify hot issues and methodological innovations in the field;
        \item \textbf{Coverage Completeness:} Good coverage of recent developments with minor gaps. Cutting-edge information coverage is comprehensive, including not only latest papers but also latest viewpoints from peers;
        \item \textbf{Source Quality:} Mostly high-quality sources with reliable attribution;
    \end{itemize}
    \vspace{0.5em}
    \textbf{Score 3: Moderate}
    \begin{itemize}[topsep=0pt, partopsep=0pt, itemsep=0pt]
        \item \textbf{Recency:} Response identifies a certain number of recent research achievements, covering some important latest developments;
        \item \textbf{Impact Level:} Includes moderately impactful research with some selection issues. Can point out some emerging trends and methodological shifts but may overlook certain key breakthroughs;
        \item \textbf{Coverage Completeness:} Adequate coverage but misses some important developments. Generally reflects the field's current state but coverage of the most cutting-edge exploratory work is insufficient;
        \item \textbf{Source Quality:} Mixed source quality with some reliability concerns;
    \end{itemize}
    \vspace{0.5em}
    \textbf{Score 2: Basic}
    \begin{itemize}[topsep=0pt, partopsep=0pt, itemsep=0pt]
        \item \textbf{Recency:} Limited recent research, misses important developments. Response identifies a small amount of recent research but misses most important latest achievements;
        \item \textbf{Impact Level:} Focuses on lower-impact or less significant research. Fails to adequately reflect the field's current active state and latest trends;
        \item \textbf{Coverage Completeness:} Poor coverage with significant gaps in recent developments. Coverage of cutting-edge developments is unsystematic, occasionally mentioning new directions but lacking complete narrative;
        \item \textbf{Source Quality:} Low-quality sources with questionable reliability;
    \end{itemize}
    \vspace{0.5em}
    \textbf{Score 1: Poor}
    \begin{itemize}[topsep=0pt, partopsep=0pt, itemsep=0pt]
        \item \textbf{Recency:} Response lacks coverage of high-impact recent work, with almost no identification of recent or cutting-edge research. Lacks recent research coverage, predominantly outdated information;
        \item \textbf{Impact Level:} No coverage of impactful or breakthrough research;
        \item \textbf{Coverage Completeness:} Severely limited coverage missing most recent developments;
        \item \textbf{Source Quality:} Description of current research state significantly differs from reality. Very poor or unreliable sources;
    \end{itemize}
\end{tcolorbox}

\begin{tcolorbox}[title=Information Coverage (Breadth), colback=gray!10, colframe=customdeepPurple!50, rounded corners, coltitle=black,
fonttitle=\bfseries, boxrule=1pt, width=\textwidth, breakable]
    \setlength{\parskip}{0pt}
    \textbf{Definition} Output should provide: (Coverage) comprehensive review of proposed focus areas, citing various representative papers, discussing the most current information from various sources, rather than just a few (1-2) papers.
\vspace{0.5em}

    \textbf{Score 5: Exceptional}
    \begin{itemize}[topsep=0pt, partopsep=0pt, itemsep=0pt]
        \item \textbf{Domain Scope:} Comprehensive coverage: answer covers various different papers and viewpoints, providing comprehensive field overview;
        \item \textbf{Perspective Diversity:} Multiple viewpoints and approaches from different research communities. Includes important discussion points not explicitly mentioned in the original question;
        \item \textbf{Methodological Range:} Covers various research methodologies and theoretical frameworks;
        \item \textbf{Interdisciplinary Connections:} Excellent integration of insights from related fields;
    \end{itemize}
    \vspace{0.5em}
    \textbf{Score 4: Strong}
    \begin{itemize}[topsep=0pt, partopsep=0pt, itemsep=0pt]
        \item \textbf{Domain Scope:} Broad coverage: output covers the field, discussing various representative papers and materials;
        \item \textbf{Perspective Diversity:} Good variety of viewpoints with most major perspectives covered. While providing broad overview, it may miss some small areas or other documents that could enhance comprehensiveness;
        \item \textbf{Methodological Range:} Covers most relevant methodological approaches;
        \item \textbf{Interdisciplinary Connections:} Good integration with some cross-field insights;
    \end{itemize}
    \vspace{0.5em}
    \textbf{Score 3: Moderate}
    \begin{itemize}[topsep=0pt, partopsep=0pt, itemsep=0pt]
        \item \textbf{Domain Scope:} Discusses representative works with satisfactory overview. Output discusses several representative works and provides satisfactory field overview;
        \item \textbf{Perspective Diversity:} Adequate variety of viewpoints but may miss some important perspectives. However, adding more papers or discussion points could significantly improve the answer;
        \item \textbf{Methodological Range:} Covers basic methodological approaches with some gaps. Covers core aspects of the question but may miss some details;
        \item \textbf{Interdisciplinary Connections:} Limited cross-field integration;
    \end{itemize}
    \vspace{0.5em}
    \textbf{Score 2: Basic}
    \begin{itemize}[topsep=0pt, partopsep=0pt, itemsep=0pt]
        \item \textbf{Domain Scope:} Partial coverage, misses important research directions. Output covers some key aspects of the field but misses important research directions, or focuses too narrowly on few sources;
        \item \textbf{Perspective Diversity:} Limited viewpoints, potential bias in selection. Lacks comprehensive perspective, failing to adequately represent field work diversity;
        \item \textbf{Methodological Range:} Narrow methodological coverage;
        \item \textbf{Interdisciplinary Connections:} Poor cross-field integration;
    \end{itemize}
    \vspace{0.5em}
    \textbf{Score 1: Pool}
    \begin{itemize}[topsep=0pt, partopsep=0pt, itemsep=0pt]
        \item \textbf{Domain Scope:} Severely limited coverage, focuses on single domain. Severely lacks coverage: output lacks coverage of several core research areas or focuses mainly on a single work area;
        \item \textbf{Perspective Diversity:} Very narrow perspective, lacks diversity. Lacking overall field perspective;
        \item \textbf{Methodological Range:} Single or very limited methodological approach;
        \item \textbf{Interdisciplinary Connections:} No cross-field integration;
    \end{itemize}
\end{tcolorbox}

\begin{tcolorbox}[title=Relevance, colback=gray!10, colframe=customdeepPurple!50, rounded corners, coltitle=black,
fonttitle=\bfseries, boxrule=1pt, width=\textwidth, breakable]
    \textbf{Definition} Evaluate whether the response stays on topic and maintains clear focus to provide useful answers to questions. Specifically, output should: 1. Adequately address core points of original question and meet your information needs (if factual). 2. Not contain much secondary information unrelated to original question.
    \setlength{\parskip}{0pt}
    \vspace{0.5em}
    
    \textbf{Score 5: Focused and entirely on topic}
    \begin{itemize}[topsep=0pt, partopsep=0pt, itemsep=0pt]
        \item \textbf{Topic Focus:} Response consistently stays closely on topic with clear focus on solving the problem;
        \item \textbf{Information Relevance:} Every piece of information directly contributes to comprehensive topic understanding;
        \item \textbf{Content Quality:} Sufficient depth of understanding and coverage of core information;
        \item \textbf{User Needs:} Fully addresses core points of original question and meets information needs;
    \end{itemize}
    \vspace{0.5em}
    \textbf{Score 4: Mostly On-Topic with Minor Deviations}
    \begin{itemize}[topsep=0pt, partopsep=0pt, itemsep=0pt]
        \item \textbf{Topic Focus:} Response is basically topic-relevant and clearly focuses on solving the problem;
        \item \textbf{Information Relevance:} Most content directly relates to the main question with minor irrelevant details;
        \item \textbf{Content Quality:} Minor off-topic deviations that temporarily distract from topic focus but don't significantly impact clarity;
        \item \textbf{User Needs:} Adequately addresses most core points with minimal distraction;
    \end{itemize}
    \vspace{0.5em}
    \textbf{Score 3: Somewhat on topic but with several digressions or irrelevant information}
    \begin{itemize}[topsep=0pt, partopsep=0pt, itemsep=0pt]
        \item \textbf{Topic Focus:} Response still revolves around original question but frequently deviates from topic;
        \item \textbf{Information Relevance:} Contains some redundant information or minor irrelevant points;
        \item \textbf{Content Quality:} Noticeable digressions that affect focus but main topic remains discernible;
        \item \textbf{User Needs:} Partially addresses core points but with unnecessary diversions;
    \end{itemize}
    \vspace{0.5em}
    \textbf{Score 2: Frequently Off-Topic with Limited Focus}
    \begin{itemize}[topsep=0pt, partopsep=0pt, itemsep=0pt]
        \item \textbf{Topic Focus:} Article somewhat addresses the question but frequently deviates from topic;
        \item \textbf{Information Relevance:} Contains significant amount of irrelevant information or unrelated points;
        \item \textbf{Content Quality:} Multiple diversions that don't help with main question and reduce overall utility;
        \item \textbf{User Needs:} Limited success in addressing core points of original question;
    \end{itemize}
    \vspace{0.5em}
    \textbf{Score 1: Off-topic}
    \begin{itemize}[topsep=0pt, partopsep=0pt, itemsep=0pt]
        \item \textbf{Topic Focus:} Content severely deviates from original question;
        \item \textbf{Information Relevance:} Difficult to discern relevance to the original question;
        \item \textbf{Content Quality:} Diverts user attention from intended topic and fails to provide useful answers;
        \item \textbf{User Needs:} Fails to address core points and does not meet information needs;
    \end{itemize}
\end{tcolorbox}

\begin{tcolorbox}[title=Information Depth, colback=gray!10, colframe=customdeepPurple!50, rounded corners, coltitle=black,
fonttitle=\bfseries, boxrule=1pt, width=\textwidth, breakable]
    \textbf{Definition} Evaluate whether the article provides sufficient information. Depth provides sufficient relevant information so readers can thoroughly understand each argument in the article.
    \setlength{\parskip}{0pt}
    \vspace{0.5em}
    
    \textbf{Score 5: Excellent Coverage and Amount (depth)}
    \begin{itemize}[topsep=0pt, partopsep=0pt, itemsep=0pt]
        \item \textbf{Detail Sufficiency:} Provides necessary and sufficient information with selective deep exploration. Can select materials requiring deep exploration for detailed discussion;
        \item \textbf{Technical Accuracy:} Highly accurate technical details with proper context;
        \item \textbf{Analytical Depth:} Deep analytical insights with sophisticated reasoning. Response provides all necessary and sufficient materials;
        \item \textbf{Contextual Understanding:} Excellent understanding of broader implications and context;
    \end{itemize}
    \vspace{0.5em}
    \textbf{Score 4: Good Coverage and Amount (depth)}
    \begin{itemize}[topsep=0pt, partopsep=0pt, itemsep=0pt]
        \item \textbf{Detail Sufficiency:} Includes most relevant information needed to understand the topic. Avoids excessive irrelevant details, but several points might benefit from deeper exploration or more specific examples;
        \item \textbf{Technical Accuracy:} Good technical accuracy with minor gaps;
        \item \textbf{Analytical Depth:} Good analytical insights with solid reasoning. Response includes most relevant information needed to understand the topic;
        \item \textbf{Contextual Understanding:} Good understanding of context and implications;
    \end{itemize}
    \vspace{0.5em}
    \textbf{Score 3: Acceptable Coverage and Amount (depth)}
    \begin{itemize}[topsep=0pt, partopsep=0pt, itemsep=0pt]
        \item \textbf{Detail Sufficiency:} Acceptable amount of relevant information, may lack some useful details;
        \item \textbf{Technical Accuracy:} Adequate technical accuracy with some inaccuracies;
        \item \textbf{Analytical Depth:} Output provides reasonable amount of relevant information, though it may lack some useful details.;
        \item \textbf{Contextual Understanding:} Basic understanding of context;
    \end{itemize}
    \vspace{0.5em}
    \textbf{Score 2: Limited Coverage and Amount (depth)}
    \begin{itemize}[topsep=0pt, partopsep=0pt, itemsep=0pt]
        \item \textbf{Detail Sufficiency:} Provides some relevant information but misses important details;
        \item \textbf{Technical Accuracy:} Poor technical accuracy with significant errors;
        \item \textbf{Analytical Depth:} Response provides some relevant information but misses important details that would aid full topic understanding.;
        \item \textbf{Contextual Understanding:} Poor understanding of broader context;
    \end{itemize}
    \vspace{0.5em}
    \textbf{Score 1: Lack of Coverage and Amount (depth)}
    \begin{itemize}[topsep=0pt, partopsep=0pt, itemsep=0pt]
        \item \textbf{Detail Sufficiency:} Lacks basic details needed for topic understanding;
        \item \textbf{Technical Accuracy:} Very poor technical accuracy with major errors;
        \item \textbf{Analytical Depth:} Output either lacks basic details needed for adequate topic understanding (e.g., method definitions, relationships between methods);
        \item \textbf{Contextual Understanding:} No understanding of context or implications;
    \end{itemize}
\end{tcolorbox}

\begin{tcolorbox}[title=Overall Helpfulness, colback=gray!10, colframe=customdeepPurple!50, rounded corners, coltitle=black,
fonttitle=\bfseries, boxrule=1pt, width=\textwidth, breakable]
    \textbf{Definition} Do you find the provided answer overall helpful? Does it assist with your literature review? Evaluate the overall utility of the response for research and learning purposes.
    \setlength{\parskip}{0pt}
    \vspace{0.5em}
    
    \textbf{Score 5: Super Useful. I can fully trust the answer}
    \begin{itemize}[topsep=0pt, partopsep=0pt, itemsep=0pt]
        \item \textbf{Question Addressing:} Answer provides comprehensive field overview and fully answers the question;
        \item \textbf{Source Quality:} Provides high-quality, trustworthy sources with comprehensive coverage;
        \item \textbf{Research Utility:} Serves as complete foundation for research without need for independent verification;
        \item \textbf{Information Reliability:} I believe I don't need to independently search for other papers or detailed information;
    \end{itemize}
    \vspace{0.5em}
    \textbf{Score 4: Useful. I may try to verify some details, but overall gives great summary}
    \begin{itemize}[topsep=0pt, partopsep=0pt, itemsep=0pt]
        \item \textbf{Question Addressing:} Answer provides detailed information and good overview of the area of interest;
        \item \textbf{Source Quality:} Provides high-quality, fresh sources across multiple sources with good diversity;
        \item \textbf{Research Utility:} Requires minimal additional editing, serves as excellent foundation for further work;
        \item \textbf{Information Reliability:} May need to check details of 1-2 specific papers/sources, but overall highly reliable;
    \end{itemize}
    \vspace{0.5em}
    \textbf{Score 3: Provides some useful discussions and papers, though requires independent reading}
    \begin{itemize}[topsep=0pt, partopsep=0pt, itemsep=0pt]
        \item \textbf{Question Addressing:} Answer is generally helpful and provides good overview with diverse perspectives;
        \item \textbf{Source Quality:} Provides at least 2-3 useful information sources previously unknown to reader;
        \item \textbf{Research Utility:} Can base further reading on recommended papers, good starting point for deeper research;
        \item \textbf{Information Reliability:} May need to independently verify some details or consult other core research papers;
    \end{itemize}
    \vspace{0.5em}
    \textbf{Score 2: Better than searching from scratch but limited utility}
    \begin{itemize}[topsep=0pt, partopsep=0pt, itemsep=0pt]
        \item \textbf{Question Addressing:} Answer provides at least one useful starting point but discussions are somewhat irrelevant;
        \item \textbf{Source Quality:} Provides at least one useful paper that can be read carefully;
        \item \textbf{Research Utility:} Limited utility for research purposes, requires significant additional work;
        \item \textbf{Information Reliability:} Overall discussions don't provide sufficiently useful information for the topic;
    \end{itemize}
    \vspace{0.5em}
    \textbf{Score 1: Unhelpful}
    \begin{itemize}[topsep=0pt, partopsep=0pt, itemsep=0pt]
        \item \textbf{Question Addressing:} Answer doesn't address the question or provides confusing information;
        \item \textbf{Source Quality:} Hasn't conducted effective retrieval, still generating using pretrained knowledge;
        \item \textbf{Research Utility:} Cannot serve as useful starting point for learning or writing relevant content;
        \item \textbf{Information Reliability:} Fails to provide understanding of literature in this field;
    \end{itemize}
\end{tcolorbox}

\subsubsection{System Design Evaluation (-2 to +2 Scale)}
\begin{table}[htbp]
\centering
\renewcommand{\arraystretch}{1.5}
\begin{tabular}{p{8cm}*{5}{>{\centering\arraybackslash}p{1cm}}}
\textbf{Evaluation Dimension} & \textbf{-2} & \textbf{-1} & \textbf{0} & \textbf{+1} & \textbf{+2} \\
\hline
\textbf{Transparency:} Decision-making process visibility & $\bigcirc$ & $\bigcirc$ & $\bigcirc$ & $\bigcirc$ & $\bigcirc$ \\
\hline
\textbf{Interruptibility:} Real-time intervention capability & $\bigcirc$ & $\bigcirc$ & $\bigcirc$ & $\bigcirc$ & $\bigcirc$ \\
\hline
\textbf{Fine-grained Interaction:} Interaction granularity level & $\bigcirc$ & $\bigcirc$ & $\bigcirc$ & $\bigcirc$ & $\bigcirc$ \\
\hline
\textbf{Inspiration:} Unexpected discoveries and insights & $\bigcirc$ & $\bigcirc$ & $\bigcirc$ & $\bigcirc$ & $\bigcirc$ \\
\hline
\textbf{Collaboration:} Collaborative partnership quality & $\bigcirc$ & $\bigcirc$ & $\bigcirc$ & $\bigcirc$ & $\bigcirc$ \\
\end{tabular}
\caption{System Design Assessment Rubric}
\end{table}

\vspace{1em}
\begin{tcolorbox}[
    title=System Design Evaluation Definition, 
    colback=gray!10, colframe=customdeepPurple!50, coltitle=black,
fonttitle=\bfseries, rounded corners, boxrule=1pt, width=\textwidth, breakable, enhanced, before skip=1em, after skip=1em
]
\setlength{\parskip}{0pt}
\textbf{Question:} Does the system design provide sufficient transparency in decision-making processes?
\vspace{0.5em}

\textbf{Interruptibility (Interruptible at any time):} To what extent do you think interruptibility can help correct the model's research approach and reduce model errors?
\vspace{0.5em}

\textbf{Fine-grained and Bidirectional Interaction:} How fine-grained do you think the current system's interaction is? (Interaction refers to nodes where users can provide input to the model)
\vspace{0.5em}

\textbf{Inspirational Perspectives (Shared cognitive context as exploration space):}
How much information in the model's decision and search process exceeded your expectations? Did it help inspire you?
\vspace{0.5em}

\textbf{Inspirational Perspectives (Shared cognitive context as exploration space):} How much information in the model's decision and search process exceeded your expectations? Did it help inspire you?
\vspace{0.5em}

\textbf{Long-term Collaboration Willingness:} Deep research systems can all interact (Deep Cognition during process, other 3 systems after research process). Research is a dynamic, multi-round complex long-term task. To what extent do these systems' interaction methods (including input methods and system feedback output methods) make you willing to engage in long-term, multi-round communication and collaboration with the system?
\vspace{0.5em}

\textbf{Long-term Collaboration Willingness:}
Deep research systems can all interact (Deep Cognition during process, other 3 systems after research process). Research is a dynamic, multi-round complex long-term task. To what extent do these systems' interaction methods (including input methods and system feedback output methods) make you willing to engage in long-term, multi-round communication and collaboration with the system?
\vspace{0.5em}

\textbf{+2 points - Excellent:}
\begin{itemize}[topsep=0pt, partopsep=0pt, itemsep=0pt]
    \item \textbf{Process Visibility:} Complete visibility of thinking, actions, and browsed content;
    \item \textbf{Decision Rationale:} Clear explanation of all decision-making processes;
    \item \textbf{Source Verification:} Full source verification and citation transparency;
    \item \textbf{Strategy Disclosure:} Complete disclosure of search and analysis strategies;
\end{itemize}

\vspace{0.5em}
\textbf{+1 points - Good:}
\begin{itemize}[topsep=0pt, partopsep=0pt, itemsep=0pt]
    \item \textbf{Process Visibility:} Good transparency with some decision process visibility;
    \item \textbf{Decision Rationale:} Adequate explanation of major decisions;
    \item \textbf{Source Verification:} Good source transparency with minor gaps;
    \item \textbf{Strategy Disclosure:} Partial disclosure of strategies and approaches;
\end{itemize}

\vspace{0.5em}
\textbf{0 points - Neutral:}
\begin{itemize}[topsep=0pt, partopsep=0pt, itemsep=0pt]
    \item \textbf{Process Visibility:} Neutral/adequate transparency level;
    \item \textbf{Decision Rationale:} Basic explanation of some decisions;
    \item \textbf{Source Verification:} Adequate source information;
    \item \textbf{Strategy Disclosure:} Limited strategy disclosure;
\end{itemize}

\vspace{0.5em}
\textbf{-1 points - Poor:}
\begin{itemize}[topsep=0pt, partopsep=0pt, itemsep=0pt]
    \item \textbf{Process Visibility:} Limited transparency, unclear decision processes;
    \item \textbf{Decision Rationale:} Poor explanation of decision-making;
    \item \textbf{Source Verification:} Limited source transparency;
    \item \textbf{Strategy Disclosure:} Minimal strategy disclosure;
\end{itemize}

\vspace{0.5em}
\textbf{-2 points - Extremely Poor:}
\begin{itemize}[topsep=0pt, partopsep=0pt, itemsep=0pt]
    \item \textbf{Process Visibility:} Black box operation with no process visibility;
    \item \textbf{Decision Rationale:} No explanation of decision-making processes;
    \item \textbf{Source Verification:} No source transparency or verification;
    \item \textbf{Strategy Disclosure:} No disclosure of strategies or methods;
\end{itemize}
\end{tcolorbox}

\vspace{1em}

\subsubsection{Deep Cognition Specific Evaluation}
\textbf{Qualitative indicator:} When comparing the Deep Cognition system with other deep research systems, do the system's functional designs (interruptibility, transparent thinking process, transparent behavioral paths, presenting search queries, displaying retrieved content) enhance this system's collaborative attributes?

\textbf{Follow-up questions: } A. If enhanced, can you provide specific examples? Which functions enhanced collaborative attributes? B. During model behavior review, could the model provide new insights/unexpected search information?

\vspace{0.5em}

\begin{table}[htbp]
\centering
\begin{tabular}{l|p{8cm}}
\textbf{Feature} & \textbf{Description} \\
\hline
Text Input & Basic text communication capability \\
\hline
Question Clarification & System's ability to clarify ambiguous queries \\
\hline
Expert Information Integration & Incorporating domain expertise \\
\hline
Thinking Process Visibility & Transparency of reasoning steps \\
\hline
Decision Process & Clarity of decision-making rationale \\
\hline
Interruptibility & Effectiveness of real-time intervention \\
\hline
Content Summary Reading & Quality of information synthesis \\
\hline
Search Query Visibility & Transparency of search strategies \\
\end{tabular}
\caption{Deep Cognition Feature-Specific Ratings (1-5 Scale)}
\end{table}

\vspace{1em}

\subsection{Post-Study-2}

\textit{Deep Cognition Evaluation: -2 for strongly negative, 0 for neutral, 2 for strongly positive}

\vspace{0.5em}
\textbf{1. Enhanced Effectiveness (Enhance cognitive efficiency or not)}

To what extent do you think this collaborative approach can improve final report generation quality (organization and consistency/information coverage/information density (depth)/relevance/overall helpfulness)?
\begin{center}
\begin{tabularx}{\textwidth}{X|c|X}

\textbf{Dimension} & \textbf{Score (-2/-1/0/1/2)} & \textbf{Reason} \\
\hline
Organization and consistency & & \\
\hline
Information coverage & & \\
\hline
Information density (depth) & & \\
\hline
Relevance & & \\
\hline
Overall helpfulness & & \\

\end{tabularx}
\end{center}
\vspace{0.5em}

\textbf{2. Results-worth-effort}
Interacting with these systems costs your time and energy. Do you think it's worth it? How worthwhile?

\begin{center}
\begin{tabularx}{\textwidth}{X|c|X}

\textbf{System} & \textbf{Score (-2/-1/0/1/2)} & \textbf{Reason} \\
\hline
Deep Cognition & & \\
\hline
OpenAI & & \\
\hline
Gemini & & \\
\hline
Grok 3 & & \\

\end{tabularx}
\end{center}
\vspace{0.5em}

\textbf{3. Research Stage Evaluation}

At which stages do you think interrupting the model's operation can effectively improve subsequent report quality? Which stage can enhance your real-time collaboration willingness with the model?

Current model nodes include: evaluating research status, generating search queries, filtering webpage URLs, browsing webpages, extracting summaries from webpages and determining usefulness, prioritizing information retrieved from webpages and organizing arguments.

You may define research stages according to your own understanding when asking this question.
\vspace{0.5em}

\textbf{Follow-up questions:}

a) At which stage of model research development is your collaboration willingness higher?

b) Can the model's research process provide you with insights? Can you give an example (screenshot or text)?

c) At which stages do you think interrupting the model's operation can more effectively improve subsequent report quality? Which stage can enhance your real-time collaboration willingness with the model?
\vspace{0.5em}

\textbf{4. Usage Willingness and Learning Cost (Interaction Willingness)}

\textbf{Quantitative indicators:} To what extent are you willing to use this system? How are the learning costs and operational burden?

\begin{center}
\begin{tabularx}{\textwidth}{X|c|X}

\textbf{Aspect} & \textbf{Score (-2/-1/0/1/2)} & \textbf{Reason} \\
\hline
Usage willingness & & \\
\hline
Ease of operation & & \\

\end{tabularx}
\end{center}
\vspace{0.5em}

\textbf{5. Feature Evaluation}

How helpful are these features for your research process? Rate (1-5) and explain reasons.

\begin{center}
\begin{tabularx}{\textwidth}{c|X|c|X}

\textbf{Feature Number} & \textbf{Feature Name} & \textbf{Score} & \textbf{Comments} \\
\hline
1 & Send text & & \\
\hline
2 & Clarify questions & & \\
\hline
3 & Add expert information & & \\
\hline
4 & Thinking process & & \\
\hline
6 & Decision & & \\
\hline
7 & Interruptible & & \\
\hline
8 & Read summaries & & \\
\hline
9 & Search queries & & \\

\end{tabularx}
\end{center}

\end{document}